\documentclass[runningheads]{llncs}
\usepackage{graphicx}

\usepackage{tikz}
\usepackage{comment}
\usepackage{amsmath,amssymb} 
\usepackage{color}
\usepackage{orcidlink}

\usepackage[accsupp]{axessibility}  

\usepackage[capitalize]{cleveref}
\crefname{section}{Sec.}{Secs.}
\Crefname{section}{Section}{Sections}
\Crefname{table}{Table}{Tables}
\crefname{table}{Tab.}{Tabs.}
\usepackage{booktabs}
\usepackage{pifont}
\newcommand{\cmark}{\ding{51}}%
\newcommand{\xmark}{\ding{55}}%
\usepackage{multirow}

\begin{document}
\pagestyle{headings}
\mainmatter
\def\ECCVSubNumber{2756}  

\title{Multimodal Transformer for Automatic 3D Annotation and Object Detection} 

\titlerunning{MTrans for Automatic Annotation}
%
\author{Chang Liu\orcidlink{0000-0001-9791-4594} \and
Xiaoyan Qian\orcidlink{0000-0003-4426-0667 } \and
Binxiao Huang\orcidlink{0000-0001-5316-703X} \and
Xiaojuan Qi\orcidlink{0000-0002-4285-1626} \and
Edmund Lam\orcidlink{0000-0001-6268-950X} \and
Siew-Chong Tan\orcidlink{0000-0001-9007-8749} \and
Ngai Wong\orcidlink{0000-0002-3026-0108}}
\authorrunning{C.Liu et al.}
%
\institute{The University of Hong Kong, Pokfulam, Hong Kong \\
\email{\{lcon7, qianxy10, huangbx7\}@connect.hku.hk}\\
\email{\{xjqi,elam,sctan,nwong\}@eee.hku.hk}}
\maketitle

\begin{abstract}
Despite a growing number of datasets being collected for training 3D object detection models, significant human effort is still required to annotate 3D boxes on LiDAR scans. To automate the annotation and facilitate the production of various customized datasets, we propose an end-to-end multimodal transformer (MTrans) autolabeler, which leverages both LiDAR scans and images to generate precise 3D box annotations from weak 2D bounding boxes. To alleviate the pervasive sparsity problem that hinders existing autolabelers, MTrans densifies the sparse point clouds by generating new 3D points based on 2D image information. With a multi-task design, MTrans segments the foreground/background, densifies LiDAR point clouds, and regresses 3D boxes simultaneously. Experimental results verify the effectiveness of the MTrans for improving the quality of the generated labels. By enriching the sparse point clouds, our method achieves 4.48\% and 4.03\% better 3D AP on KITTI moderate and hard samples, respectively, versus the state-of-the-art autolabeler. MTrans can also be extended to improve the accuracy for 3D object detection, resulting in a remarkable 89.45\% AP on KITTI hard samples. Codes are at \url{https://github.com/Cliu2/MTrans}. 

\keywords{
3D Autolabeler, 3D Object Detection, Multimodal Vision, Self-attention, Self-supervision, Transformer

}
\end{abstract}

\section{Introduction}
\label{sec:intro}
3D detection technologies have seen rapid development in recent years, with considerable importance for tasks such as autonomous driving and robotics. Large-scale datasets can often improve the performance and generality of deep neural networks, hence new datasets such as nuScenes~\cite{caesar2020nuscenes} and Waymo Open Dataset~\cite{sun2020scalability_waymo} include millions of annotated objects and hundreds of thousands of LiDAR scans. Nonetheless, the annotation procedure is extremely labor-intensive, especially for 3D point clouds~\cite{meng2020weakly,wei2021fgr}. Therefore, it is imperative to explore ways to accelerate the annotation procedure.

\begin{figure}
    \centering
    \includegraphics[width=0.8\textwidth]{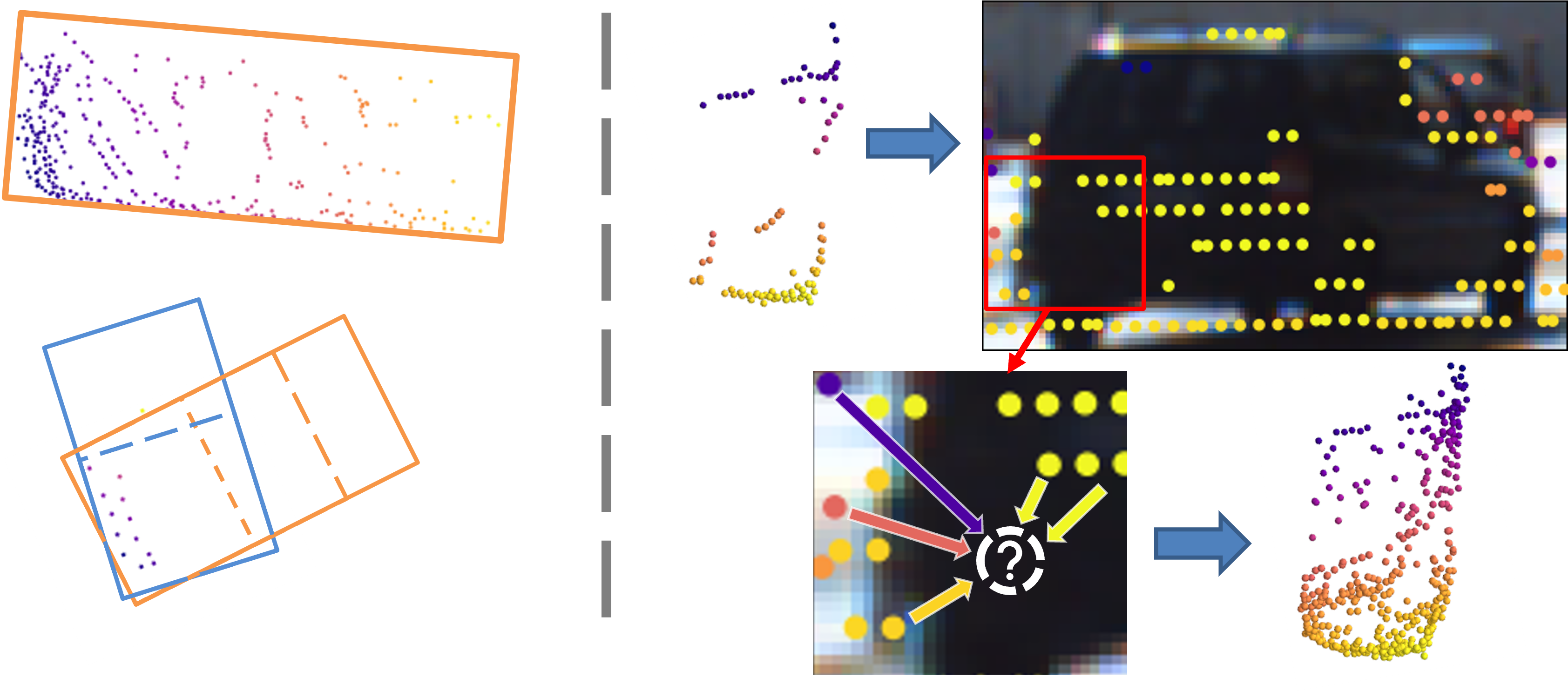}
    \caption{\textbf{Left:} The sparsity issue of point cloud for 3D box annotation. Compared with the top object, the bottom is ambiguous in orientation and scale. Both are from the Bird's Eye View of LiDAR scans. \textbf{Right:} Interpolating 3D points by referring to context points based on 2D semantic and geometric relationships. Point's depth shown in color.}
    \label{fig:sparsity_issue}
\end{figure}

Compared with 3D annotations, 2D bounding boxes are much easier to obtain. Hence, we investigate the generation of 3D annotations from weak 2D boxes. Despite its substantial practical value, only a few existing works study the automatic annotation problem. Taking point clouds and images as the multimodal inputs, FGR~\cite{wei2021fgr} converts 2D boxes to 3D boxes with a non-learning algorithm. Alternatively, WS3D~\cite{meng2021ws3d_towards} leverages center clicks as weak annotations, and directly regresses 3D boxes from cylindrical proposals centered by the clicks. Both methods leverage the image information to reduce the problem space and achieves state-of-the-art performances on the KITTI dataset~\cite{Geiger2012CVPR_kitti}. Nevertheless, their methods overlook the pervasive sparsity issue of point clouds. As shown on the left of~\cref{fig:sparsity_issue}, even though the problem domain is narrowed down by the weak 2D annotations, it is still challenging to generate a 3D box if the points are too sparse, due to the ambiguity in orientation and scale. Consequently, although these methods generate human-comparable annotations for easy samples, they suffer obvious accuracy drops for hard samples that are usually sparse, by 9.13\% and 6.47\% in 3D AP, respectively.

The sparsity problem remains a fundamental challenge for automatic annotations. Besides the sparsity issue, point clouds lack color information. Therefore, similar-shaped objects (e.g., traffic light post and pedestrian) are difficult to distinguish, which further complicates the problem~\cite{vora2020pointpainting}. Fortunately, images provide dense fine-grid RGB information, which is an effective complement to the sparse point clouds. Motivated by this, we investigate point cloud enrichment with image information. As shown on the right of \cref{fig:sparsity_issue}, 3D LiDAR points can be projected onto the image, and therefore an image pixel's 3D coordinates can be estimated by referring to the context LiDAR points, based on their 2D spatial and semantic relationships. This way, new 3D points can be generated from 2D pixels, and hence densify the sparse point cloud.

To this end, we propose the end-to-end multimodal transformer (MTrans) which alleviates the sparsity problem and outputs accurate 3D box annotations given weak 2D boxes. In order to leverage the dense image information to enrich sparse point clouds, we present the multimodal self-attention module, which efficiently extracts multimodal 2D and 3D features, modeling points' geometric and semantic relationships. To train MTrans, we also introduce a multi-task design, so that it learns to generate new 3D points to reduce the sparsity of point clouds while predicting 3D boxes. The multi-task design also includes a foreground segmentation task to encourage MTrans to uncover point-wise semantics. The multimodal self-attention and multi-task design work collectively and enable the MTrans to efficiently exploit multimodal information to address the sparsity issue, and hence facilitate more accurate 3D box prediction. Moreover, to exploit unlabelled data for the point generation task, a \emph{mask-and-predict} self-supervision strategy can be carried out. With a small amount of annotated data (e.g., 500 frames) and optional unlabeled data, our method can generate high-quality 3D box annotations for training other 3D object detectors. Comprehensive experiments are conducted, and our MTrans outperforms all the other baseline methods significantly. Using only 500 annotated frames, PointRCNN trained with MTrans can achieve 99.62\% of its original performance trained with full human annotations, which is 3.65\% higher in mAP than the state-of-the-art autolabeler. Furthermore, our point generation approach for handling the sparsity issue also has direct value for 3D object detection tasks. Extended for object detection, MTrans surpasses similar 2D-box-aided methods of F-PointNet~\cite{qi2018frustumpnet} and F-ConvNet~\cite{wang2019frustum-convnet} by 19.6\% and 10.8\% mAP, respectively. Our contributions are threefold: 

\begin{enumerate}
    \item We propose a novel autolabeler, MTrans, which generates high-quality 3D boxes from weak 2D bounding boxes, thus greatly saving manual work.
    \item MTrans effectively alleviates the pervasive sparsity issue for 3D point clouds, and it can also be extended for 3D object detection tasks. 
    \item Comprehensive experiments are conducted to justify our method. MTrans outperforms existing state-of-the-art autolabelers by large margins. Visualization results are also provided for interpretability.
\end{enumerate}

\section{Related Work}

\subsection{Multimodal 3D Object Detection}
Several previous studies investigate multimodal networks, utilizing images and LiDAR point clouds for a more comprehensive perception. Most of them extract features from images and point clouds in different branches, followed by fusing the information for final decision~\cite{chen2017mv3d,huang2020epnet,zhao2021ACMNet,xie2020pircnn}. MV3D~\cite{chen2017mv3d} proposes the deep fusion manner to repetitively exchange information from the different views including the image, LiDAR's front view and Bird's Eye View (BEV), during the inference. Later, EPNet~\cite{huang2020epnet} adopts Feature Pyramid Network to encode LiDAR's front view and RGB images, and fuses the two feature pyramids with the gating mechanism. Similarly, ACMNet~\cite{zhao2021ACMNet} and 3D-CVF~\cite{yoo20203d-cvf} are based on gating fusion as well. PI-RCNN~\cite{xie2020pircnn} presents a hybrid PACF fusion module. To align different views of the image and point clouds during fusion, AVOD~\cite{ku2018AVOD} introduces the 3D anchor grid to generate 3D proposals and applies fusion only for the proposed region. Alternatively, Pi-RCNN~\cite{xie2020pircnn} and PointPainting~\cite{vora2020pointpainting} directly extract point cloud features with point-based backbones (e.g., PointNet~\cite{qi2017pointnet} and PointNet++~\cite{qi2017pointnet++}), and augment them with image features through LiDAR-image calibration.

Rather than fusing different modalities, some other research employs images to narrow down the problem space. PointPainting~\cite{vora2020pointpainting} performs image semantic segmentation and highlights the points corresponding to the interested 2D region. Similarly, Ref.~\cite{mccraith2021lifting2d} performs instance segmentation on images to filter the 3D point clouds. F-PointNet~\cite{qi2018frustumpnet} and F-Convnet~\cite{wang2019frustum-convnet} leverage 2D bounding boxes on images to locate a frustum region of the point cloud, and hence lowers the difficulty and improves accuracy. CLOCs~\cite{pang2020clocs} generates 2D box proposals from images to filter 3D proposals, exploiting the consistency of the two modalities to improve the overall detection accuracy.

The approaches above demonstrated the value of multiple modalities. The dense image data can complement the sparse point clouds. Nevertheless, the number of points remains unchanged, hence the sparsity problem is not substantially solved. Also, the multi-branch design introduces extra computation and memory overheads, leaving it still an open question on how to fully utilize multimodal information~\cite{wang2020what_makes}. 

\subsection{Single-modal Detectors Addressing the Sparsity Problem}
Noticing the sparsity problem, some researchers try to enrich the sparse clouds to improve detection accuracy. SPG~\cite{xu2021spg} firstly locates foreground regions, where an object is likely to exist, and then generates semantic points with a voxel-based point generation module. Specifically, the network predicts whether each voxel is in the foreground or background, and then generates a new point for each foreground voxel. BtcDet~\cite{xu2021behind} densifies the point clouds within voxel-based proposals as well. Instead of directly generating new points, BtcDet finds a best matched prior object for each proposal, and densifies the proposal by mirroring the original points as well as aggregating the points from the prior object.

Both SPG and BtcDet alleviate the sparsity problem, resulting in state-of-the-art performances on the KITTI detection dataset. The two works demonstrate the value of enriching the sparse point cloud for the 3D object detection task. Nonetheless, they are all trained with a large amount of annotated data, which are not accessible for the automatic annotation task. Additionally, they only leverage single modality, omitting the dense RGB information from images.

\subsection{Autolabelers for 3D Object Detection}
Although autolabelers have considerable potential for real-world applications, only a few works have investigated this problem. FGR~\cite{wei2021fgr} is a non-learning approach that takes 2D bounding boxes to locate frustum sub-point-clouds and removes ground points with the RANSAC algorithm. After that, they heuristically calculate the tightest 3D bounding box that can wrap all remaining points for each object. In contrast, other deep-learning approaches train their networks with a small amount of human annotations, and generate 3D boxes from inexpensive weak annotations.

CC~\cite{tang2019cc_semi} extracts frustum sub-clouds in the same way as FGR, but adopts PointNets~\cite{qi2017pointnet} for segmenting and regressing 3D boxes. SDF~\cite{zakharov2020sdf} uses predefined CAD models to estimate the 3D geometry of cars detected in 2D images. VS3D~\cite{qin2020vs3d} generates 3D proposals based on cloud density with an unsupervised UPM module. Points in the proposals together with cropped images are fed into a classification/regression network for the final 3D boxes. Later, WS3D~\cite{meng2020weakly} and WS3D(2021)~\cite{meng2021ws3d_towards} utilize center clicks as weak 2D annotations. Human annotators are asked to click on the  objects' centroids on images, followed by adjusting the position in BEV of LiDAR scans. Points within a cylindrical proposal centered by each click are processed by a backbone network for the 3D pseudo box.

The above autolabelers utilize image modality and weak annotations to automatically generate 3D boxes for point clouds. However, they mainly use images to coarsely locate objects but do not address the sparsity problem of point clouds. Consequently, their performances are still sub-optimal compared with human annotations, having noticeable accuracy drops, especially for moderate and hard samples, which are usually far and sparse.

\section{MTrans for Automatic Annotation}
\label{sec:method_autolabel}
In this section, we first define the automatic annotation problem, and then introduce the proposed MTrans as an autolabeler, including multimodal self-attention mechanism, multi-task design, and self-supervised point generation. 

\subsubsection{Automatic Annotation.} Given the LiDAR scan and its corresponding image, this research aims to generate 3D bounding box annotations for interested objects (i.e., Cars) from weak annotations of 2D bounding boxes. LiDAR points and image pixels are connected by the LiDAR-image projection based on the calibration parameters. Each point in the LiDAR point cloud can be projected to a pixel on the corresponding image. The autolabeler is trained with a small amount of human annotated data (e.g., 500 frames) and then used to label the remaining data. Another object detection network can be trained with the auto-generated 3D labels, saving manual work significantly. 

\begin{figure}[ht]
    \centering
    \includegraphics[width=0.95\textwidth]{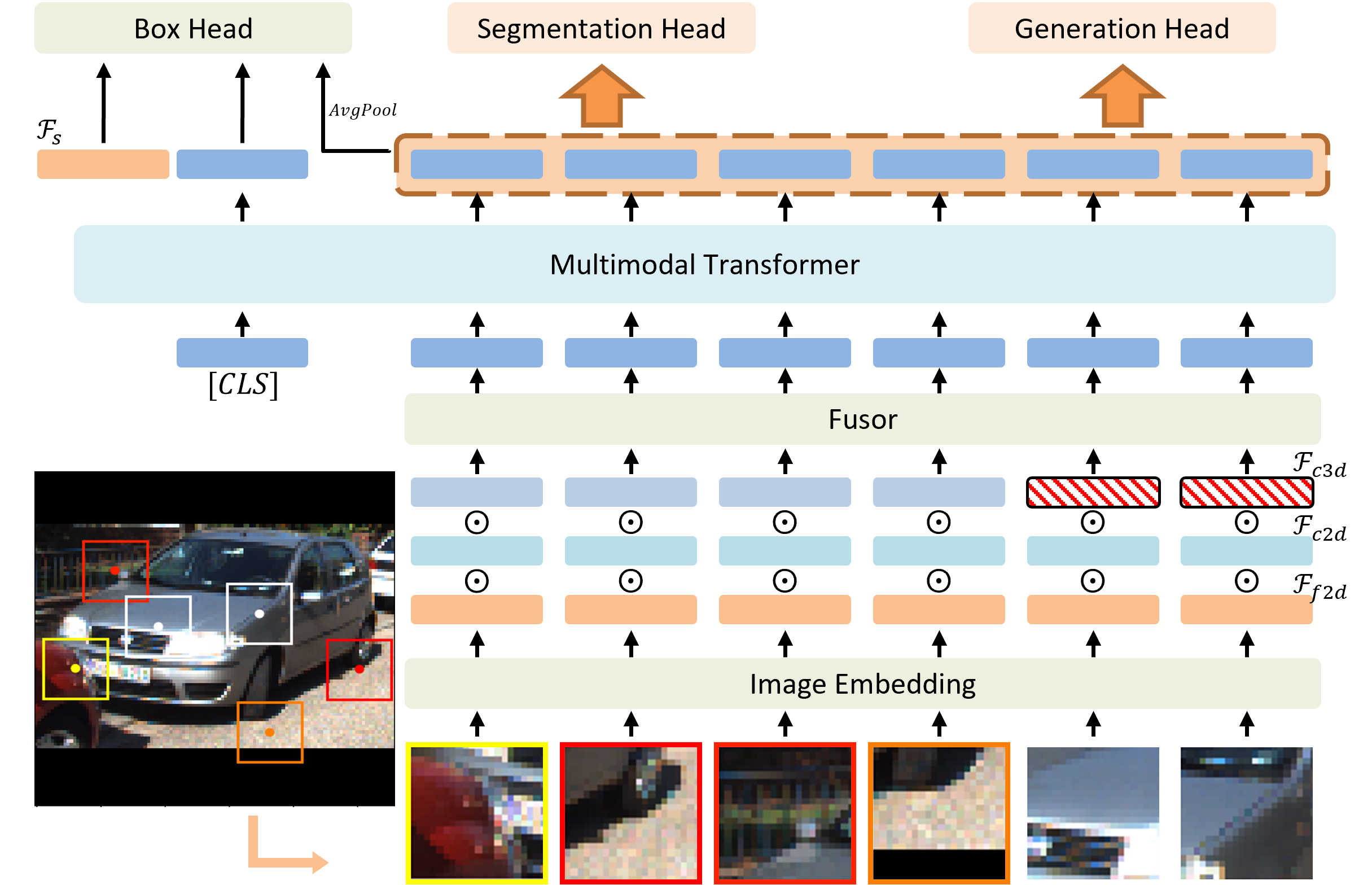}
    \caption{Workflow of the MTrans. The network models both 3D and 2D information from the LiDAR scans and images. The frustum points and the image are fed into the network for simultaneous foreground segmentation, point generation, and box regression. $\mathcal{F}_{f2d}$, $\mathcal{F}_{c2d}$ and $\mathcal{F}_{c3d}$ are the embeddings for image patches, 2D coordinates and 3D coordinates. $\mathcal{F}_{s}$ is the global image feature derived by averaging $\mathcal{F}_{f2d}$.}
    \label{fig:1_workflow}
\end{figure}

\subsubsection{Overview.} As shown in \cref{fig:1_workflow}, to address the sparsity problem and generate high quality 3D annotations, our MTrans takes multimodal inputs of LiDAR point clouds and their corresponding images (\cref{sec:data_preparation}). The multimodal inputs are encoded and fused as point-level embedding vectors for further processing (\cref{sec:embedding_layers}). With the proposed multimodal self-attention module, object features are extracted based on the points' semantic and geometric relationships (\cref{sec:self_attention_layer}). Then the extracted features are used for multiple tasks of foreground segmentation, point generation, and 3D box regression, encouraging the MTrans to comprehensively utilize the multimodal information and alleviate the sparsity problem (\cref{sec:prediction_head}). Additionally, a self-supervised training strategy is proposed to leverage unlabeled data for the point generation task (\cref{sec:self-supervise}).

\subsection{Data Preparation}
\label{sec:data_preparation}
Our MTrans generates one 3D bounding box for each object in a weak 2D box. Using the LiDAR-image calibration parameters, a 3D point $(x, y, z)$ can be projected onto the image plane $(u, v)$ by the mapping function $f_{cal}$. Therefore, we can extract the frustum sub-cloud $\mathcal{P}_{F}$ corresponding to a 2D box:
\begin{equation}
    \mathcal{P}_{F} = \{(x, y, z)\; |\; f_{cal}(x, y, z) \in \mathcal{B}_{2D}\}, 
\end{equation}
where $\mathcal{B}_{2D}$ is the region cropped by the 2D box. We use $n$ to represent the frustum cloud size (i.e., the number of points). The points' 2D projections are defined as:
\begin{equation}
    \mathcal{C}_{2D} = \{(u, v)\; |\; (u, v) = f_{cal}(x, y, z), \; (x, y, z) \in \mathcal{P}_{F} \}, 
\end{equation}
A shown in \cref{fig:1_workflow}, centered by the 2D projections, image patches of shape $k\times k$ are extracted, denoted as $\mathcal{I}_{p}$. Thus, every LiDAR point has three categories of features, $\mathcal{P}_{F}^{(i)}$, $\mathcal{C}_{2D}^{(i)}$ and $\mathcal{I}_{p}^{(i)}$.

As mentioned in~\cref{sec:intro}, in order to alleviate the sparsity problem, we generate 3D points from the image pixels, by referring to the $n$ real LiDAR points as \emph{context}. Since the number of real LiDAR points, $n$, varies from sample to sample, we randomly drop points or sample pixels from images as \emph{padding}, resulting in fixed $n'$ elements. Besides the padding points, we further sample another $m$ pixels uniformly from the image, called \emph{target} points. The padding and target points only have 2D features ($\mathcal{C}_{2D}^{(i)}$ and $\mathcal{I}_{p}^{(i)}$) without known 3D coordinates ($\mathcal{P}_{F}^{(i)}$). 

\subsection{Feature Embeddings and Object Representation}
\label{sec:embedding_layers}
Same as vanilla transformers~\cite{vaswani2017transformer,devlin2018bert}, we encode the object features as embedding vectors. For the 2D position, we use sinusoidal position embedding~\cite{vaswani2017transformer} to encode $u$ and $v$ into a half-length vector respectively, and then concatenate them for the 2D position embedding $\mathcal{F}_{c2d}$. As the 3D coordinates are continuous, a multilayer perceptron (MLP) is used to encode the $(x, y, z)$ coordinates into 3D position embeddings $\mathcal{F}_{c3d}$. Lastly, we extract image patch features $\mathcal{F}_{f2d}$ with a 3-layer convolutional sub-network from $\mathcal{I}_{p}$. Specifically, we empirically set the patch size as $7\times7$ and kernel size as $3\times3$. All of $\mathcal{F}_{c2d}$, $\mathcal{F}_{c3d}$ and $\mathcal{F}_{f2d}$ have shapes of $((n'+m), d)$, where $d$ is the embedding vector length, for the $n'+m$ points. Note that for the padding and target points, they have no known 3D coordinates. Therefore, their $\mathcal{F}_{c3d}$ features are replaced by a trainable embedding vector $\mathcal{E}$.

Afterwards, we fuse the three kinds of features by concatenating them along the $d$-axis, and decrease the channel from $3d$ back to $d$ by a fully-connected layer. As shown in \cref{fig:1_workflow}, we introduce an object-level token $[CLS]$, which is a randomly-initialized trainable embedding vector. The total $n'+m+1$ element representations of the object, $\mathcal{F} \in \mathbb{R}^{(n'+m+1)\times d}$, are then input into the MTrans for feature extraction with the proposed multimodal self-attention mechanism.

\begin{figure}[ht]
    \centering
    \includegraphics[width=0.8\textwidth]{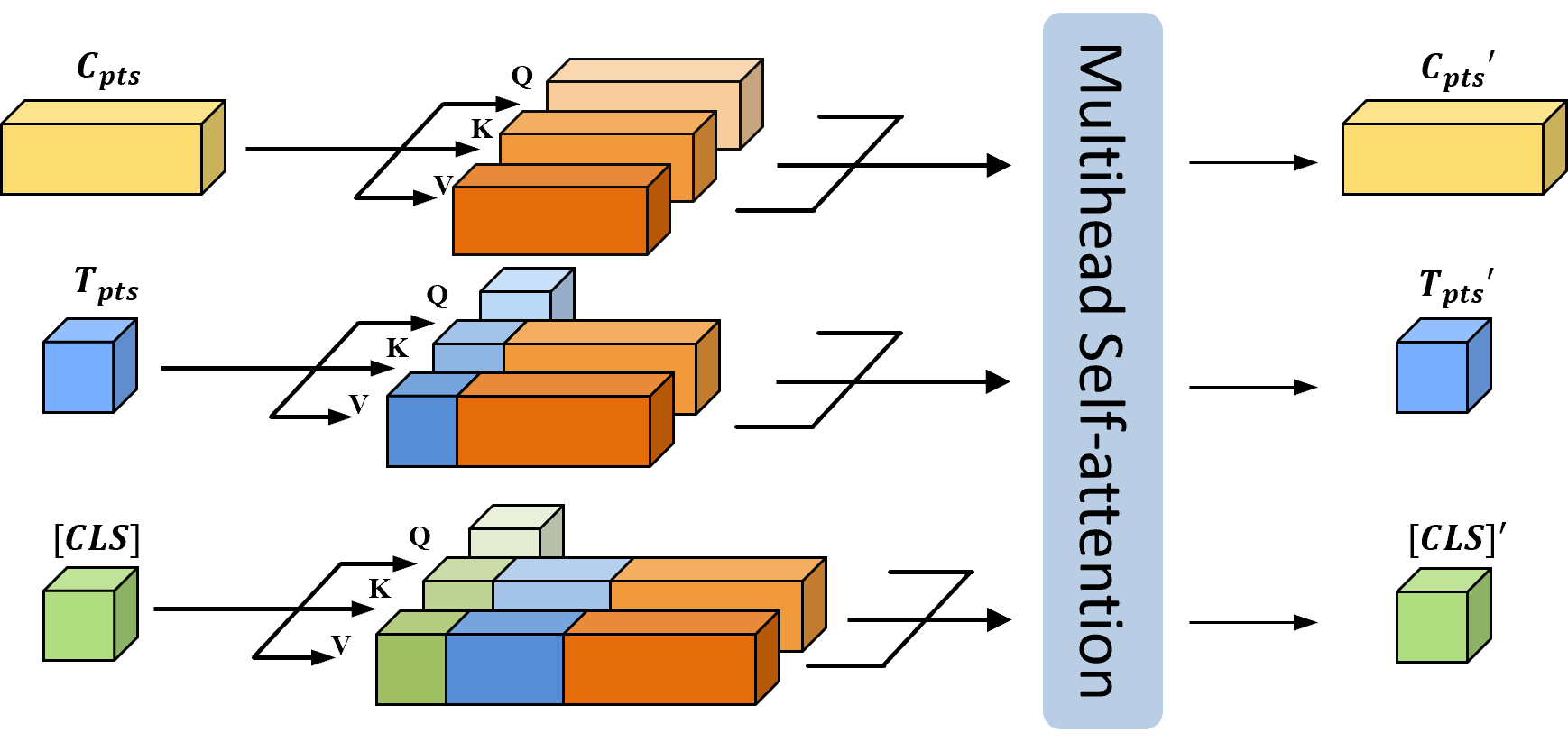}
    \caption{Multimodal Self-attention. C-pts with known 3D coordinates attend to each other, while each of the T-pts attends to itself as well as the C-pts. $[CLS]$ attends to all available C-pts and T-pts for object-level representation. For better visual effect, we show one T-pt doing self-attention in the middle row. Best viewed in color. }
    \label{fig:f2_self_attn}
\end{figure}
\subsection{Multimodal Self-attention}
\label{sec:self_attention_layer}
Following the vanilla transformer~\cite{vaswani2017transformer,devlin2018bert}, we manipulate the element representations with the multi-head self-attention mechanism. The element representations are transformed into query $Q$, key $K$, and value $V$. The values $V$ are recombined based on attention scores calculated by multiplying $Q$ and $K$. Owing to the space limitation, we do not repeat the technical details of the multi-head self-attention mechanism, but only stress on the modifications for our MTrans.

Different from the original transformer, under our problem setting, some elements contain incomplete information (i.e., the padding and target points have only 2D information but no 3D information). Therefore, we treat the original frustum sub-cloud points with full information as the context points (\emph{C-pts} in \cref{fig:f2_self_attn}), and recover 3D coordinates for the padding and target elements (\emph{T-pts}). Specifically, the T-pts look up C-pts based on their position and semantic relationships, and interpolate their 3D coordinates to enrich the sparse point cloud. During our multimodal self-attention, the C-pts only attend to each other but not T-pts, while each T-pt can also attend to itself, and $[CLS]$ attends to all available points to aggregate object-level information. 

As shown in \cref{fig:f2_self_attn}, the C-pts perform vanilla self-attention, and have representations updated as $\text{C-pts}'$. For each T-pts, a one-element $Q$ is built from its element representation, while $K$ and $V$ are from the concatenation of itself and the C-pts. Similarly, $[CLS]$ has $Q$ from itself, but $K$ and $V$ from the concatenation of $[CLS]$, T-pts, and C-pts. With this asymmetric design, T-pts update themselves by absorbing 3D information from C-pts, and $[CLS]$ encodes a global representation of the object. For our MTrans, we stack 4 multimodal self-attention layers, and vanilla feed-forward-layer~\cite{vaswani2017transformer} is adopted between each pair of the self-attention layers.

\subsection{Multi-task Prediction Heads}
\label{sec:prediction_head}
After the self-attention layers, the transformed element representations are fed into multiple prediction heads for the multi-task supervision. Firstly, a 2-layer MLP $\mathcal{H}_{seg}$ performs binary segmentation for each point. Points within the ground-truth 3D bounding box are regarded as foreground, otherwise background. The segmentation loss $\mathcal{L}_{seg}$ is the summation of Cross-Entropy loss and Dice loss. Since T-pts have no 3D coordinates, only C-pts are supervised.

Another 2-layer MLP head $\mathcal{H}_{xyz}$ predicts the 3D coordinates for the point generation task. Since there are no ground-truth values for the padding and target points, we adopt a mask-and-predict strategy to train this task with self-supervision, which will be introduced in detail in the next section. Smooth $L_1$ loss is used for the point generation task as $\mathcal{L}_{xyz}$. Intuitively, the foreground is more critical than background, so we only calculate $\mathcal{L}_{xyz}$ for the foreground points predicted by $\mathcal{H}_{seg}$. Moreover, we multiply the loss with a weight of 0.1 if the point is a context point due to the leaked 3D coordinates, otherwise 1.

The third prediction head $\mathcal{H}_{box}$ directly regresses 3D bounding boxes from the object-level representations, which is derived by concatenating the $[CLS]$ representation, the averaged element representations, and the averaged image feature of $\mathcal{F}_{s}$, as shown in \cref{fig:1_workflow}. The output box is a 7-element vector for the location ($(x, y, z)$ coordinates), dimension (length, width, height), and rotation along the z-axis. We use the dIoU loss~\cite{iouloss,pytorchiou,zheng2020distanceiou} for the box loss $\mathcal{L}_{box}$.

As the IoU (intersection over union) metric is direction-invariant (i.e., rotating a box for 180 degrees does not change the IoU), we further introduce a direction head for the binary direction classification. Specifically, orientation within $[-\pi/2, \pi/2)$ is regarded as the front, and $[-\pi, -\pi/2) \cup [\pi/2, \pi]$ as the back. Cross Entropy loss is used for the direction loss $\mathcal{L}_{dir}$.

All the heads are trained simultaneously, and the overall loss is given by: 
\begin{equation}
    \mathcal{L} = \mathcal{L}_{seg} + \mathcal{L}_{xyz}+\lambda_{box}\mathcal{L}_{box} + \mathcal{L}_{dir},
\end{equation}
where the $\lambda_{box}$ is a coefficient and we empirically set it as 5.

\subsection{Self-supervised Point Generation}
\label{sec:self-supervise}
One major motivation of the proposed MTrans is to enrich the sparse point clouds with image information. As mentioned above, beyond the context points, some extra T-pts (i.e., padding and target points) are appended with unknown 3D information. The point generation task, supervised by the loss $\mathcal{L}_{xyz}$, aims to uncover the 3D coordinates for these T-pts based on their 2D geometric and semantic relationships with the context points. Due to the lack of ground truths, we adopt a self-supervised strategy for this task. Specifically, we randomly mask a portion ($0\sim95\%$) of the context points, by replacing their $\mathcal{F}_{c3d}$ with the trainable embedding vector $\mathcal{E}$. The masked points are then treated as T-pts and supervised by the $\mathcal{L}_{xyz}$. With this partial supervision, we managed to train the MTrans for the point generation task. During the inference, no context points are masked and the model recovers the 3D coordinates for the real T-pts.

\section{Extension of MTrans to 3D Object Detection}

\label{sec:detection_with_mt}
Inspired by F-PointNet~\cite{qi2018frustumpnet} and F-ConvNet~\cite{wang2019frustum-convnet}, we can also convert the MTrans into a 3D object detector. Similarly, our model lifts 2D detection results up to 3D bounding boxes. The architecture and workflow remain the same as in \cref{sec:method_autolabel}. However, to be compatible for the detection metric of Average Precision (AP), we add another prediction head $\mathcal{H}_{c}$ for the confidence of the generated 3D box. The confidence is indicated by the estimated IoU score of each generated 3D box. Instead of using a small amount of data for the automatic annotation task, the detector version of the MTrans is trained with full data. The overall loss is also updated as:
\begin{equation}
    \mathcal{L} = \mathcal{L}_{seg} + \mathcal{L}_{xyz}+\lambda_{box}\mathcal{L}_{iou} + \mathcal{L}_{dir} + \mathcal{L}_{c},
\end{equation}
where the new term $\mathcal{L}_{c}$ is the Smooth $L_1$ loss for confidence head $\mathcal{H}_c$.

\section{Experimental Evaluations}
\label{sec:experiment}
The proposed MTrans is evaluated for two tasks, the automatic annotation and 3D object detection. We also conduct ablation studies to justify the contributions of each module.

\subsubsection{KITTI Dataset.} KITTI~\cite{Geiger2012CVPR_kitti} is a benchmark dataset for 3D detection for autonomous driving. We follow the official practice to split the dataset into training and validation sets with 3,712 and 3,769 frames. The standard evaluation metric of AP with the IoU threshold being 0.7 is adopted. Same as FGR~\cite{wei2021fgr}, we focus on the Car class and filter out objects with less than 5 foreground LiDAR points, due to some samples having too few points to possibly draw a 3D box.

\subsection{Implementation Details}
\label{sec:implementation}
For the MTrans, 4 multimodal self-attention layers with the hidden size of 768 and 12 heads are used. As an autolabeler, the model is trained with 500 or 125 annotated frames for 300 epochs. The remaining frames are also used as unlabeled data for self-supervision on the point generation task. Adam optimizer is employed with a learning rate of $10^{-4}$. We also use standard image augmentation methods of auto contrast, random sharpness and color jittering, as well as point cloud augmentation methods of random translation, scaling and mirroring. Due to some 2D boxes having overlaps, an overlap mask of shape is added as a new channel of the image, where the overlapped areas have values of 0, and otherwise 1. For each object, the cloud size $n'$ is set to 512 for context plus padding points, and $m$ is 784 for the number of image-sampled target points.

\begin{table}[ht]
    \centering
    \caption{Automatic annotation results on KITTI \emph{val} set and \emph{test} set.}
    \setlength{\tabcolsep}{4pt}
    \resizebox{\linewidth}{!}{
    \begin{tabular}{lccccccc}
    \toprule
    \multirow{2}{*}{Method} & \multirow{2}{1.5cm}{\centering Fully Supervised} & \multicolumn{3}{c}{$\text{AP}_{3D}$ Val} & \multicolumn{3}{c}{$\text{AP}_{3D}$ Test} \\
    \cmidrule(lr){3-5} \cmidrule(lr){6-8} & & Easy & Mod. & Hard & Easy & Mod. & Hard\\
    \midrule
    PointPillars~\cite{lang2019pointpillars} & \cmark & 86.10 & 76.58 & 72.79 & 82.58 & 74.31 & 68.99 \\
    PointPillars~\cite{lang2019pointpillars} & 500f & 80.65 & 67.65 & 66.14 & - & - & -\\
    PointPillars~\cite{lang2019pointpillars} & 125f & 71.36 & 58.29 & 55.11 & - & - & -\\
    \textbf{Ours + PointPillars} & 500f & 86.69 & 76.56 & 72.38 & 77.65 & 67.48 & 62.38 \\
    \textbf{Ours + PointPillars} & 125f & 83.70 & 71.66 & 66.67 & - & - & -\\
    \midrule
    PointRCNN~\cite{shi2019pointrcnn} & \cmark & 88.99 & 78.71 & 78.21 & 86.96 & 75.64 & 70.70 \\
    PointRCNN~\cite{shi2019pointrcnn} & 500f & 88.54 & 76.85 & 70.12 & - & - & -\\
    PointRCNN~\cite{shi2019pointrcnn} & 125f & 85.63 & 73.60 & 67.98 & - & - & -\\
    \textbf{Ours + PointRCNN} & 500f & 88.72 & 78.84 & 77.43 & 83.42 & 75.07 & 68.26 \\
    \textbf{Ours + PointRCNN} & 125f & 87.64 & 77.31 & 74.32 & - & - & -\\
    \midrule
    \midrule
    \multicolumn{8}{c}{\textit{Compare with other autolabelers}} \\
    \midrule
    WS3D~\cite{meng2020weakly} & BEV Centroid & 84.04 & 75.10 & 73.29 & 80.15 & 69.64 & 63.71 \\
    WS3D(2021)~\cite{meng2021ws3d_towards} & BEV Centroid & 85.04 & \textit{75.94} & \textit{74.38} & \textit{80.99} & \textit{70.59} & \textit{64.23} \\
    FGR~\cite{wei2021fgr} & 2D Box & \textit{86.67} & 73.55 & 67.90 & 80.26 & 68.47 & 61.57 \\
    \midrule
    \textbf{Ours (PointRCNN)} & 2D Box & \textbf{88.72} & \textbf{78.84} & \textbf{77.43} & \textbf{83.42} & \textbf{75.07} & \textbf{68.26}\\
    \bottomrule
    \end{tabular}
    }
    \label{tab:kitti_result}
\end{table}

\subsection{Automatic Annotation Results} To evaluate our MTrans as an autolabeler, we employ two popular backbone object detection networks, PointRCNN~\cite{shi2019pointrcnn} and PointPillars~\cite{shi2020pvrcnn}, and retrain them with the auto-generated 3D annotations. To examine the influence of different data scales, we also train them with all frames, 500 frames, and 125 frames of human-annotated data, respectively. As shown in \cref{tab:kitti_result}, the accuracy drops dramatically when the data are insufficient, especially for the moderate and hard samples. With 125 frames of annotated data, PointPillars and PointRCNN suffer mean accuracy drops of 16.9\% and 6.23\% in AP.

Meanwhile, the proposed MTrans is also trained with 500 frames and 125 frames of annotated data, and then re-annotates the whole KITTI \emph{train} set. As shown in \cref{tab:kitti_result}, trained with the MTrans-generated annotations, the two networks both significantly outperform their counterparts. For PointPillars, our method brings 7.06\% and 12.42\% absolute average increases in AP on the \emph{val} set for 500 and 125 frames of data, respectively. Impressively, using only 500 and 125 frames, PointRCNN trained with MTrans can achieve 99.62\% and 97.24\% of the original model trained with full data on the \emph{val} set. On the \emph{test} set, MTrans-trained PointRCNN also yields 97.19\% of the original model. The above results demonstrate that our MTrans can generate high-quality 3D annotations, producing object detection networks comparable to their counterparts trained with human annotations. Automatically raising 2D boxes into 3D annotations, our method greatly saves human efforts for the annotation procedure.

We also compare our method with existing state-of-the-art autolabelers, FGR~\cite{wei2021fgr} and WS3D~\cite{meng2021ws3d_towards,meng2020weakly}. Although they also generate 3D boxes from weak 2D annotations, we want to stress the difference in task design for the three approaches. FGR uses 2D boxes, same as in MTrans, but is a non-learning algorithm, while WS3D uses center-clicks on the image and BEV, which might provide extra 3D information. Using the same PointRCNN backbone, our method surpasses all the above baselines significantly. Especially for the moderate and hard samples, MTrans improves the 3D AP by 4.48\% and 4.03\% respectively. This observation aligns with the motivation of enriching the sparse point cloud for higher quality of generated labels, since moderate and hard samples usually suffer more from the sparsity issue than easy samples.

\begin{table}[ht]
    \centering
    \caption{Detection results on KITTI \emph{val} set.}
    \setlength{\tabcolsep}{3pt}
    \resizebox{\linewidth}{!}{
    \begin{tabular}{lcccccccc}
    \toprule
    \multirow{2}{*}{Method} & \multirow{2}{*}{Modality} & \multirow{2}{1.5cm}{\centering{Extra Ref. Signal}} &  \multicolumn{3}{c}{$\text{AP}_{3D}$} & \multicolumn{3}{c}{$\text{AP}_{3D}$ R40} \\
    \cmidrule(lr){4-6} \cmidrule(lr){7-9} & & & Easy & Mod. & Hard & Easy & Mod. & Hard\\
    \midrule
    MV3D~\cite{chen2017mv3d} & LiDAR+RGB & \xmark & 71.29 & 62.68 & 56.56 & - & - & - \\
    F-PointNet~\cite{qi2018frustumpnet} & LiDAR+RGB & 2D box & 83.76 & 70.92 & 63.65 & - & - & - \\
    F-ConvNet~\cite{wang2019frustum-convnet} & LiDAR+RGB & 2D box & 89.02 & 78.80 & 77.09 & - & - & -\\ 
    VPFNet~\cite{zhu2021vpfnet} & LiDAR+RGB & \xmark & - & - & - & 93.42 & 88.76 & 86.05\\
    \midrule
    SECOND~\cite{yan2018second} & LiDAR & \xmark & 87.43 & 76.48 & 69.10 & 90.97 & 79.94 & 77.09\\
    Voxel-RCNN~\cite{deng2021voxelrcnn} & LiDAR & \xmark & 89.41 & 84.52 & 78.93 & 92.38 & 85.29 & 82.86 \\
    PV-RCNN~\cite{shi2020pvrcnn} & LiDAR & \xmark & 89.35 & 83.69 & 78.70 & 92.57 & 84.83 & 82.69 \\
    SPG~\cite{xu2021spg} & LiDAR & \xmark & 89.81 & 84.45 & 79.14 & 92.53 & 85.31 & 82.82 \\
    BtcDet~\cite{xu2021behind} & LiDAR & \xmark & - & - & - & 93.15 & 86.28 & 83.86 \\
    \midrule
    \textbf{Ours} & LiDAR+RGB & 2D box & \textbf{97.82} & \textbf{89.89} & \textbf{89.45} & \textbf{98.83} & \textbf{93.57} & \textbf{90.95}\\
    \bottomrule
    \end{tabular}
    }
    \label{tab:detection_results}
\end{table}

\subsection{Detection Results}
As mentioned in \cref{sec:detection_with_mt}, with an extra prediction head for the box confidence, the proposed MTrans can also be applied for 3D object detection task. Same as F-PointNet~\cite{qi2018frustumpnet} and F-ConvNet~\cite{wang2019frustum-convnet}, we train our MTrans with full data, and use ground-truth 2D boxes for frustum proposals. We evaluate our method on the KITTI \emph{val} set, using metrics of $\text{AP}_{3D}$ and $\text{AP}_{3D}$ R40 (40 recall thresholds).

Shown in \cref{tab:detection_results}, our method achieves impressive performances of 92.38\% and 94.46\% for mAP and mAP(R40). The detection accuracy is significantly higher than other state-of-the-art baselines. However, same as F-PointNet~\cite{qi2018frustumpnet} and F-ConvNet~\cite{wang2019frustum-convnet}, our method utilizes the ground-truth 2D boxes as extra reference signals, which reduces the problem difficulty. Nonetheless, in the fair comparison with F-PointNet and F-ConvNet under identical task and experiment settings, our method still boosts the mAP by 19.61\% and 10.75\%, respectively.

\begin{table}[ht]
    \centering
    \caption{Ablation Results.}
    \setlength{\tabcolsep}{3pt}
    \begin{tabular}{lccccccccc}
    \toprule 
    \multirow{2}{*}{} & \multirow{2}{*}{MM} & \multirow{2}{*}{Seg} & \multirow{2}{*}{Gen} & \multirow{2}{*}{ExtraPts} & \multirow{2}{*}{SelfSup} & \multicolumn{4}{c}{Metric} \\
    \cmidrule(lr){7-10} & & & & & & mIoU & Recall & mAP & $\text{mAP}_{R40}$\\
    \midrule
    Vanilla & \xmark & \xmark & \xmark & \xmark & \xmark & 60.07 & 41.53 & 48.57 & 46.89 \\
    \midrule
    Multimodal & \cmark & \xmark & \xmark & \xmark & \xmark & 65.57 & 53.30 & 58.59 & 57.33 \\
    Seg Loss & \cmark & \cmark & \xmark & \xmark & \xmark & 69.45 & 61.53 & 67.84 & 69.42 \\
    XYZ Loss & \cmark & \xmark & \cmark & \xmark & \xmark & 67.61 & 57.76 & 63.62 & 63.82 \\
    Self-supervise &\cmark & \xmark & \cmark & \xmark & \cmark & 68.42 & 59.82 & 65.25 & 66.89 \\
    Densify & \cmark & \xmark & \cmark & \cmark & \xmark & 75.37 & 74.95 & 80.82 & 83.47 \\
    Full Generate & \cmark & \xmark & \cmark & \cmark & \cmark & 76.84 & 78.48 & 83.98 & 85.72 \\
    \midrule
    Ours & \cmark & \cmark & \cmark & \cmark & \cmark & \textbf{77.50} & \textbf{81.22} & \textbf{86.24} & \textbf{90.30} \\ 
    Ours(all frames) & \cmark & \cmark & \cmark & \cmark & \cmark & 81.01 & 87.65 & 92.36 & 94.43 \\
    \bottomrule
    \end{tabular}
    \label{tab:ablation}
\end{table}
\subsection{Ablation Study}
Ablation studies are conducted to verify the technical contributions of the proposed method. We train the MTrans with 500 annotated frames of the KITTI \emph{train} set, and evaluate the model on the \emph{val} set. The generated 3D boxes are compared with the ground-truth human annotations, and three metrics are employed, namely, the mean IoU (mIoU), recall with IoU threshold of 0.7, and mean average precision (mAP, $\text{mAP}_{R40}$) over the Car class. 

As shown in \cref{tab:ablation}, the vanilla transformer, which directly processes 3D LiDAR points $\mathcal{F}_{c3d}$, cannot model the problem well with the small amount of data with only 500 frames. The \emph{Multimodal} variant introduces extra modality of images, and improves the mIoU and recall significantly by 5.5\% and 11.77\%, respectively. \emph{Seg Loss} and \emph{XYZ Loss} introduce the multi-task design, and again improve the accuracy significantly by 12.09 and 6.49\% $\text{mAP}_{R40}$, demonstrating the contribution of the auxiliary losses. For \emph{Densify}, extra target points are evenly sampled from the image, and their 3D coordinates are estimated to densify the point cloud a step further. Alternatively, \emph{Self-supervise} utilize unlabeled data for the point generation task, also improving the accuracy, resulting in 68.42\% mIoU. Combining all the point cloud densification techniques, \emph{Full Generate}'s mIoU and recall both surpass 70\%. Leveraging all the modules, our MTrans archives impressive 81.22\% Recall and 86.24\% mAP using only 500 frames of annotated data, which is even comparable with state-of-the-art object detectors trained with full annotations.

The ablation studies justify the contributions of our four innovations, namely, the multimodal network, auxiliary multiple tasks, point generation, and self-supervision. The multimodal inputs can effectively boost accuracy with negligible modification on the vanilla transformer (\emph{Multimodal} variant). Moreover, the introduction of the auxiliary multi-task design (\emph{Seg Loss} and \emph{XYZ Loss}) raises the mIoU by 2.04\% and 3.88\%, showing that the model is encouraged to understand the problem more comprehensively from multiple aspects. The largest gain, however, comes from adding extra target points from the image, which increases the mIoU by 7.76\% and mAP by 20.35\% (\emph{Densify}). This observation well supports our motivation that image-guided interpolation can enrich the sparse point cloud for better annotations. Combining all the techniques, our final MTrans yields a surprising 77.5\% mIoU and 81.22\% recall, with only 500 annotated frames. Nonetheless, the gains from these techniques tend to be saturated when combining all of the modules. Compared with the mAP accuracy for all-data-trained MTrans at the bottom of \cref{tab:kitti_result}, we hypothesize that the main constraint is the amount of training data.

\begin{figure}
    \centering
    \includegraphics[width=\textwidth]{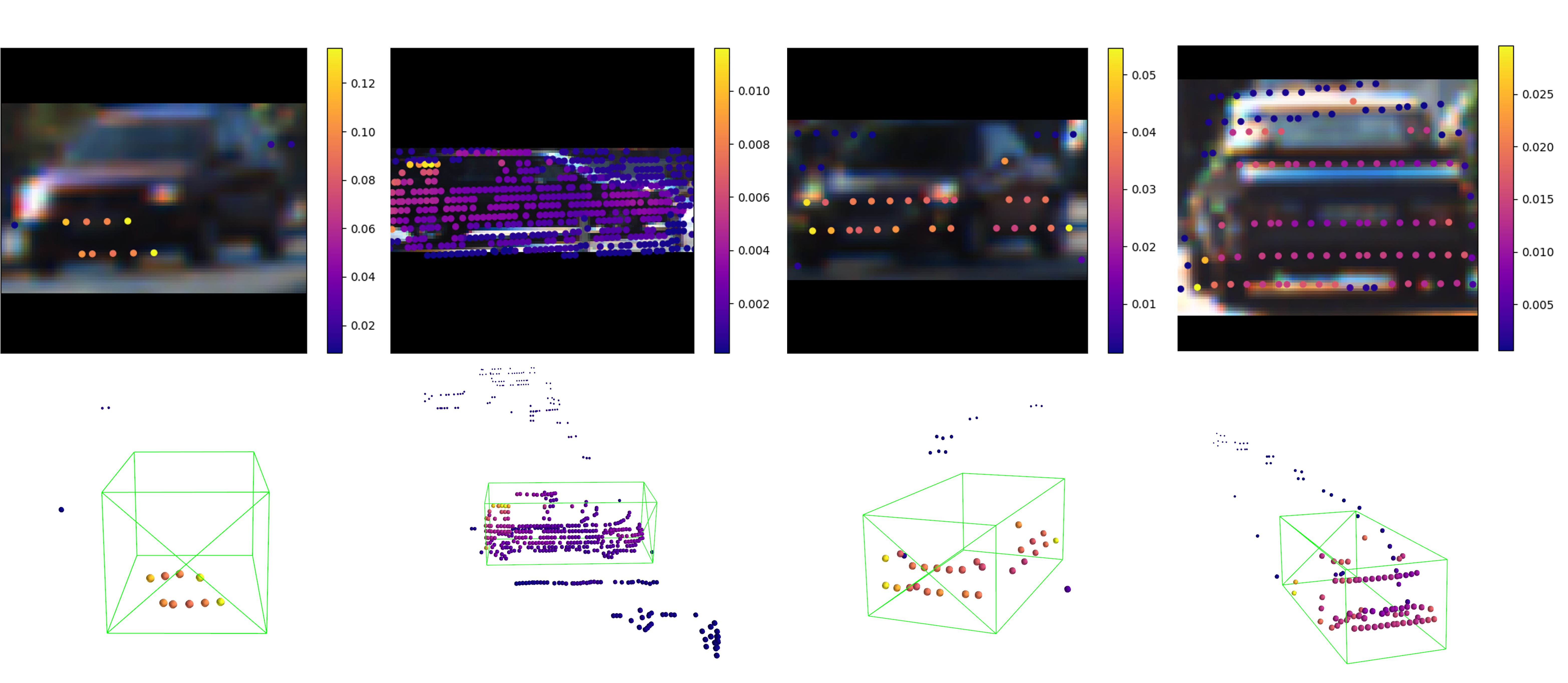}
    \caption{Attention score visualization results. Context points and their average attention scores are plotted. Each point's color represents the amount of attention it receives from other points. The attention scores sum to 1 for each data sample. It is observed that the foreground points tend to receive more attention than background points, which provides an intuitive interpretation for the multimodal attention MTrans.}
    \vspace{-5pt}
    \label{fig:vis_results}
\end{figure}
\section{Visualization of Attention on Context Points}
In \cref{fig:vis_results} we visualize the attention distribution over context points. Specifically, for each context point, we average the attention scores it received from all other points. The attention scores of all context points sum to 1. As shown in the figure, points that belong to the car tends to receive more attention than background points, which aligns with our intuition, since background points provide little reference value of labeling 3D boxes. Moreover, the attention distribution also demonstrates a coherent spatial pattern in both the 2D image contents and the 3D space. This observation suggests that multimodal self-attention extracts object features based on 3D \& 2D geometric, and 2D semantic relationships.

Another interesting phenomenon is that the foreground points closer to object edges tend to receive more attention. As shown in the figures, yellow points are presented near box corners for all four samples. This is also reasonable since corner points usually provide more critical information for 3D box prediction. The observation provides good interpretability for our MTrans.

\section{Conclusion}
This work has proposed the MTrans, an autolabeler that generates 3D boxes from weak 2D box annotations, given the LiDAR point clouds and corresponding images. Our method effectively alleviates the pervasive sparsity problem of point clouds. Specifically, we introduce the multimodal self-attention mechanism, together with auxiliary multi-task and self-supervision design, which leverage image information to generate extra 3D points. Experimental results demonstrate that our method can outperform existing autolabelers significantly, generating 3D box annotations comparable to human annotations. Moreover, the proposed MTrans can also be applied for the 3D object detection task, achieving state-of-the-art performances. Beyond LiDAR + monocular image, an interesting future direction would be to incorporate more modalities to improve the detection accuracy and robustness, such as radar and/or stereo images. It would also be valuable to investigate the domain transfer problem for the autolabelers, which could save human effort further. We hope this study can inspire future works to solve the sparsity problem as well as to develop automatic annotation workflows to reduce manual workloads.

\clearpage
%
%
\bibliographystyle{splncs04}
\bibliography{ref.bib}
\end{document}


\pagestyle{headings}
\mainmatter
\def\ECCVSubNumber{2756}  

\title{Supplementary Material for ``Multimodal Transformer for Automatic 3D Annotation and Object Detection''} 

\titlerunning{Abbreviated paper title}
%
\author{Chang Liu\orcidlink{0000-0001-9791-4594} \and
Xiaoyan Qian  \and
Binxiao Huang\orcidlink{0000-0001-5316-703X} \and
Xiaojuan Qi\orcidlink{0000-0002-4285-1626} \and
Edmund Lam\orcidlink{0000-0001-6268-950X} \and
Siew-Chong Tan\orcidlink{0000-0001-9007-8749} \and
Ngai Wong\orcidlink{0000-0002-3026-0108}}
\authorrunning{C.Liu et al.}
%
\institute{The University of Hong Kong, Pokfulam, Hong Kong \\
\email{\{lcon7, qianxy10, huangbx7\}@connect.hku.hk}\\
\email{\{xjqi,elam,sctan,nwong\}@eee.hku.hk}}
\maketitle

\section{Background of Transformer and Self-attention}
In this section, we introduce the Transformer architecture and the self-attention mechanism, which are firstly proposed by Ref.~\cite{vaswani2017transformer}. The transformer is originally an encoder-decoder network for NLP (natural language processing) tasks, which consists of a stack of self-attention layers and feedforward layers. Given sequential input data, Transformers update the hidden states of each element within the sequence by the self-attention mechanism:

$$
\begin{aligned}
&\text{Attention}(Q, K, V) = softmax(\frac{QK^T}{\sqrt{d_k}})V
\end{aligned}
$$

\noindent where the $Q, K, V$ are derived from linear transformations from the hidden states, and $d_k$ is the dimension of hidden states for scaling. By this operation, the hidden states are updated by selectively aggregating the context, weighed by the derived attention scores.

\section{nuScenes Autolabeler with MTrans} 
Besides the KITTI dataset, we also evaluate MTrans on the automatic annotation task with the nuScenes dataset~\cite{caesar2020nuscenes} to verify its generality. In this section, we train MTrans with 500 frames of annotated data and then use it to re-label the whole training set. Another object detector, PointPillars~\cite{lang2019pointpillars}, is then trained with the auto-generated labels. We also compare the generated 3D boxes with human annotated ones directly.

\subsubsection{nuScenes Dataset.} The nuScenes dataset is one of the largest open datasets for 3D detection in the field of autonomous driving, including 28,130 frames for training, 6,019 for validation and 6,008 for testing. According to the official challenge settings, similar classes and rare classes are removed, resulting in 10 out of 23 annotated classes for detection. Different from KITTI, nuScenes adopts the evaluation metric of NDS (nuScenes detection score). Moreover, the mAP is calculated by considering the 2D center distance on the ground plane instead of IoU as in KITTI. Each LiDAR frame (360 degrees) corresponds to 6 images captured by individual cameras for different view angles.

Same as in Sec.~5 in the main paper, we preprocess the dataset to extract objects with more than 5 foreground points. nuScenes does not provide 2D boxes, so we project the 3D boxes on the images, and take the outermost corners to build 2D boxes. For each 2D box, the frustum point cloud and its 2D projection are extracted. They are then fed into MTrans for estimating the 3D box. Since this research mainly focuses on the \emph{Car} class, other classes such as pedestrian are excluded.

\subsubsection{Implementation Details.} Due to nuScenes samples being sparser than KITTI ones, we empirically set the cloud size $n'$ as 350. Additionally, since we project 3D boxes on images, the overlap mask introduced in Sec.~5.1 in the main paper is inaccurate (e.g., some neighbor objects are occluded and hardly visible in the image but still results in a large overlapping region of the 2D box). Therefore, we modify the overlap masks to contain five levels, based on the visibility attribute of nuScenes objects. Consequently, the five possible mask values are set as 1 (no overlap) or 0.75, 0.5, 0.25, 0 denoting overlap with another object that has visibility level of 0-40\%, 40-60\%, 60-80\% or 80-100\%, respectively.

Again, 500 LiDAR frames are used to train our MTrans. After training, we use MTrans to re-label the whole training set for the \emph{Car} class. Because the evaluation metric of nuScenes involves all the 10 classes, we keep the human annotations for other classes but replace \emph{Car} annotations with the MTrans-generated ones. The popular object detector PointPillars~\cite{lang2019pointpillars} is re-implemented and trained with the auto-annotated data.

\begin{table}[ht]
    \centering
    \caption{Automatic annotation results on nuScenes \emph{val} set. H.Anno for human annotations. Our method achieves comparable accuracy as the original model trained with human annotations.}
    \setlength{\tabcolsep}{3pt}
    \resizebox{\linewidth}{!}{
    \begin{tabular}{lcccccccc}
        \toprule
        Method & H.Anno & mATE$\downarrow$ & mASE$\downarrow$ & mAOE$\downarrow$ & mAVE$\downarrow$ & mAAE$\downarrow$ & mAP$\uparrow$ & NDS$\uparrow$ \\
        \midrule
        PointPillars & \cmark & 35.98 & 26.27 & 37.27 & 31.57 & 19.82 & 41.26 & 55.54 \\
        \textbf{Ours} &  500f & 38.07 & 26.56 & 43.78 & 35.75 & 19.74 & 33.74 & 50.48 \\
        \bottomrule
    \end{tabular}
    }
    \label{tab:nuscenes_result}
\end{table}

\begin{table}[ht]
    \centering
    \caption{MTrans generated 3D boxes in comparison with human annotations.}
    \setlength{\tabcolsep}{4pt}
    \begin{tabular}{lcccc}
        \toprule
        Method & mIoU$\uparrow$ & Recall($\text{IoU}=0.7$)$\uparrow$ & LE$\downarrow$ & Recall($\text{LE}=0.5$)$\uparrow$\\
        \midrule
        \textbf{Ours (train)} & 74.25 & 74.94 & 0.3479 & 88.13 \\
        \textbf{Ours (val)} & 73.91 & 73.62 & 0.3448 & 88.18 \\
        \bottomrule
    \end{tabular}
    \label{tab:nuscenes_miou_result}
\end{table}

\subsubsection{Experiment Results.} As shown in~\cref{tab:nuscenes_result}, compared with the original PointPillars trained with all frames with human annotations, our method requires only 500 annotated frames yet achieves comparable performance. The NDS score of MTrans is 90.89\% of the original PointPillars that are trained with tens of thousands of human-annotated frames. In addition to the mAP and NDS, another 5 official metrics are also reported, namely, mATE (mean Average Translation Error), mASE (mean Average Scale Error), mAOE (mean Average Orientation Error), mAVE (mean Average Velocity Error) and mAAE (mean Average Attribute Error). Except for mAOE, our MTrans yields similar performance for all other error terms, where the increases are less than 1\%. Considering the enormous scale of the nuScenes dataset, MTrans can significantly save human workload for the annotation process. Moreover, the results also demonstrate that MTrans can yield good performances on different datasets, such as KITTI and nuScenes, proving the generality of our method. 

To decouple the influences of the detector network and other classes such as pedestrian, besides training another object detector with the MTrans-generated annotations, we also directly compare MTrans \emph{Car} annotations with manually labeled ground-truths. Results are shown in~\cref{tab:nuscenes_miou_result}. Trained with 500 frames, MTrans re-labels the whole training set and validation set of nuScenes. We adopt four metrics, namely the mean IoU, recall with Iou threshold of 0.7, mean location error (LE) and recall with LE threshold of 0.5.

Although no existing autolabelers have studied the dataset of nuScenes for comparsion, we find our MTrans still performs robustly on the large-scale dataset. Compared with ground-truth human annotations, 3D boxes generated by MTrans have 74.25\% mIoU and 74.94\% recall, which is close to its performances on the KITTI dataset. Moreover, since nuScenes employs location error instead of IoU for calculating mAP, it is worth noticing that the recalls with LE threshold of 0.5 are over 88\% on both the \emph{train} and \emph{val} sets. The results demonstrate that MTrans is able to generate high-quality 3D box annotations on different datasets robustly, therefore accelerating the annotation procedure for a broad range of applications.

\section{Qualitative Results}

We also show the qualitative results of MTrans, including the generated pseudo labels and densified sparse point clouds. The results demonstrate that MTrans is able to effectively address the sparsity issue and produces high-quality 3D boxes.

\subsection{Comparison with SOTA Method}
\begin{figure*}[t]
    \begin{center}
  \includegraphics[width=0.24\linewidth]{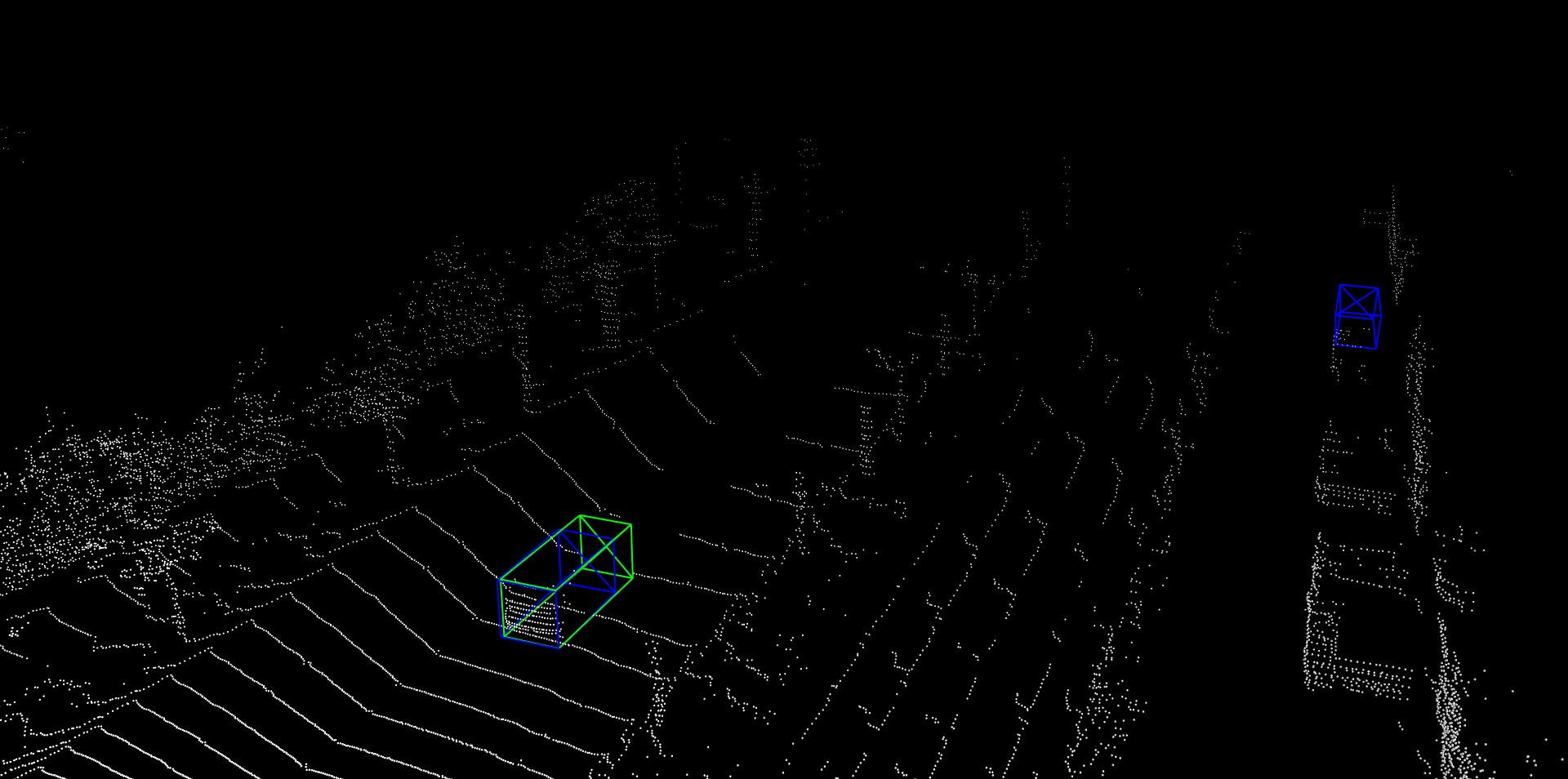} 
  \includegraphics[width=0.24\linewidth]{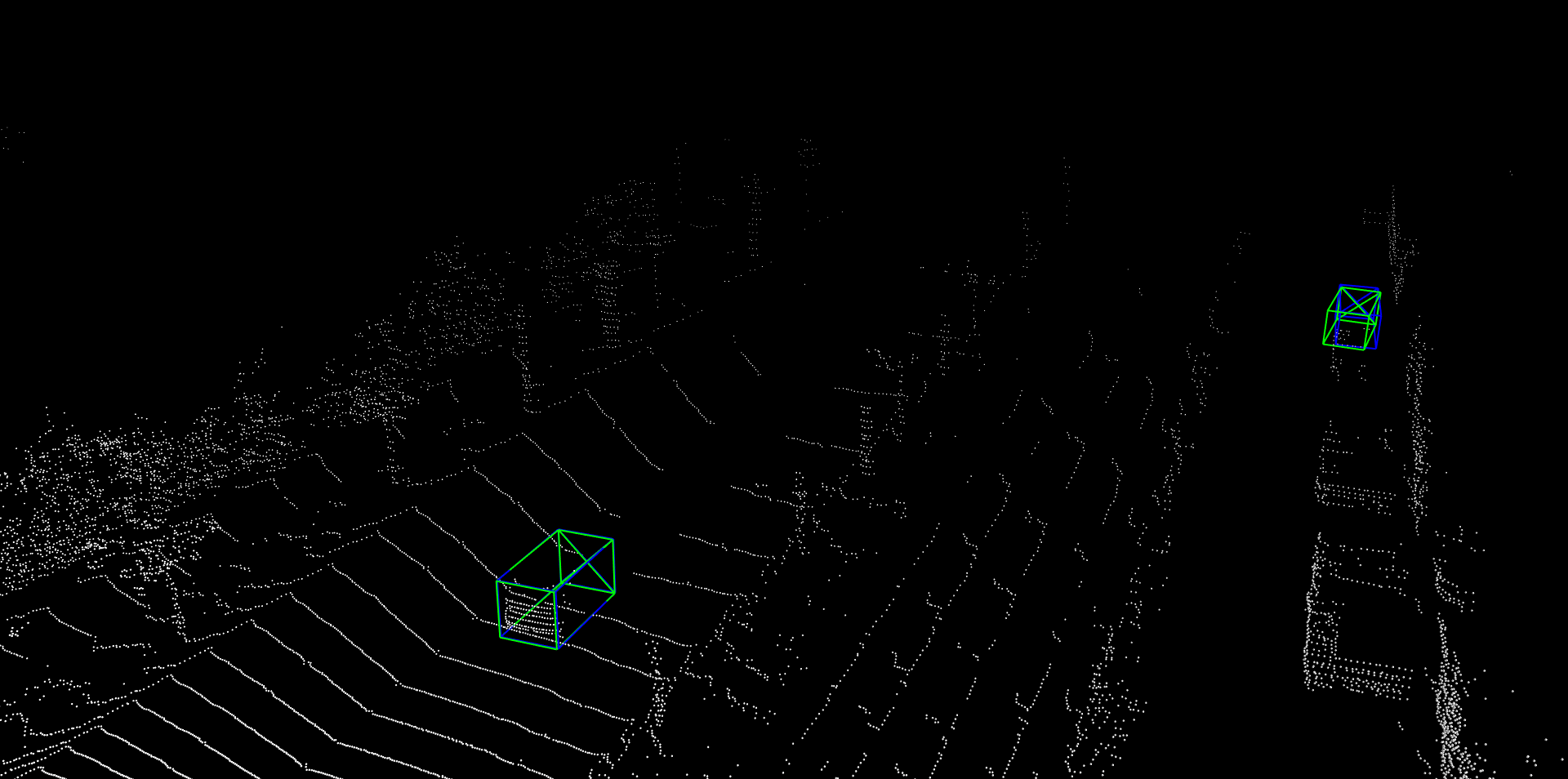}  
  \includegraphics[width=0.24\linewidth]{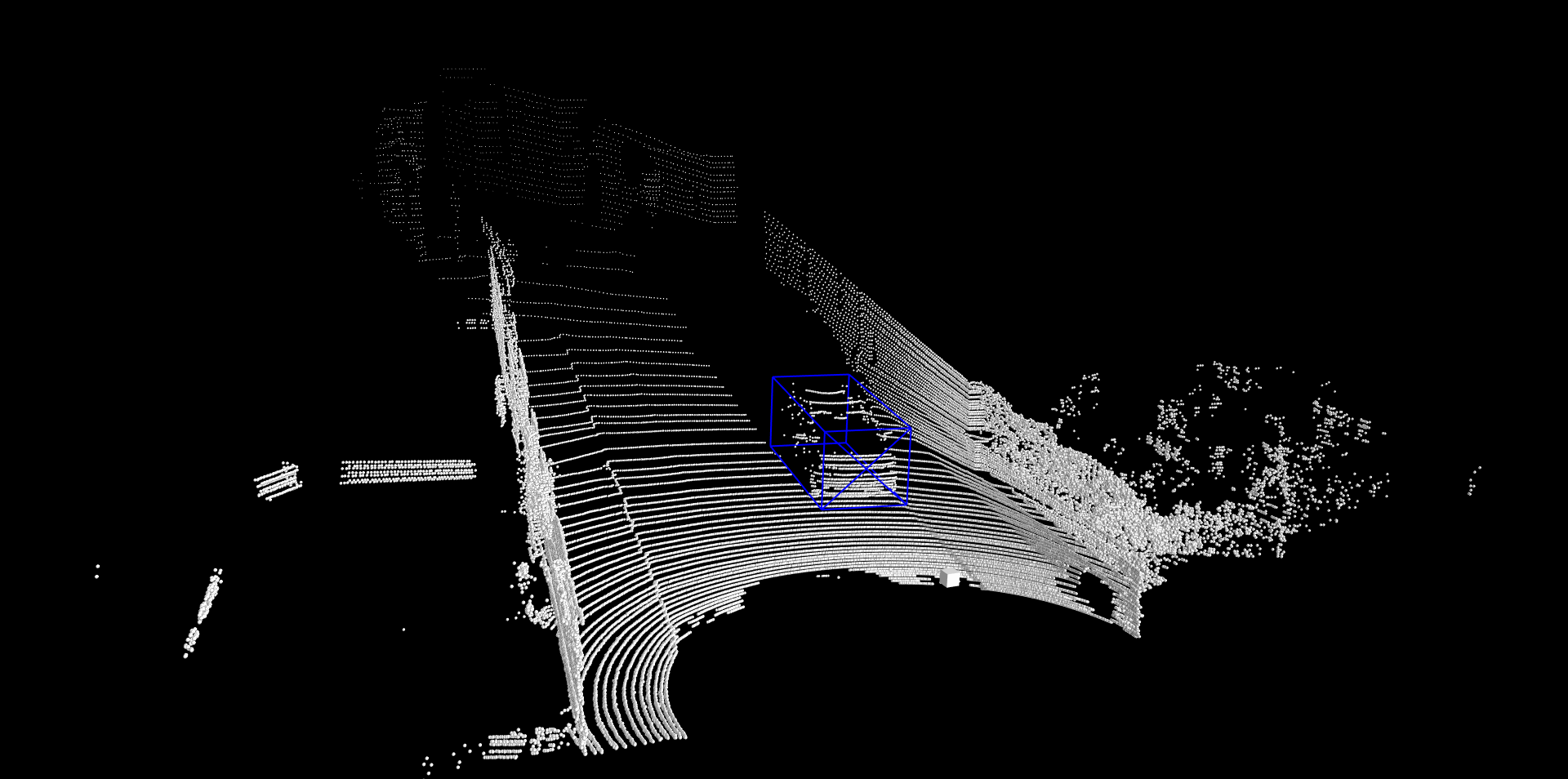} 
  \includegraphics[width=0.24\linewidth]{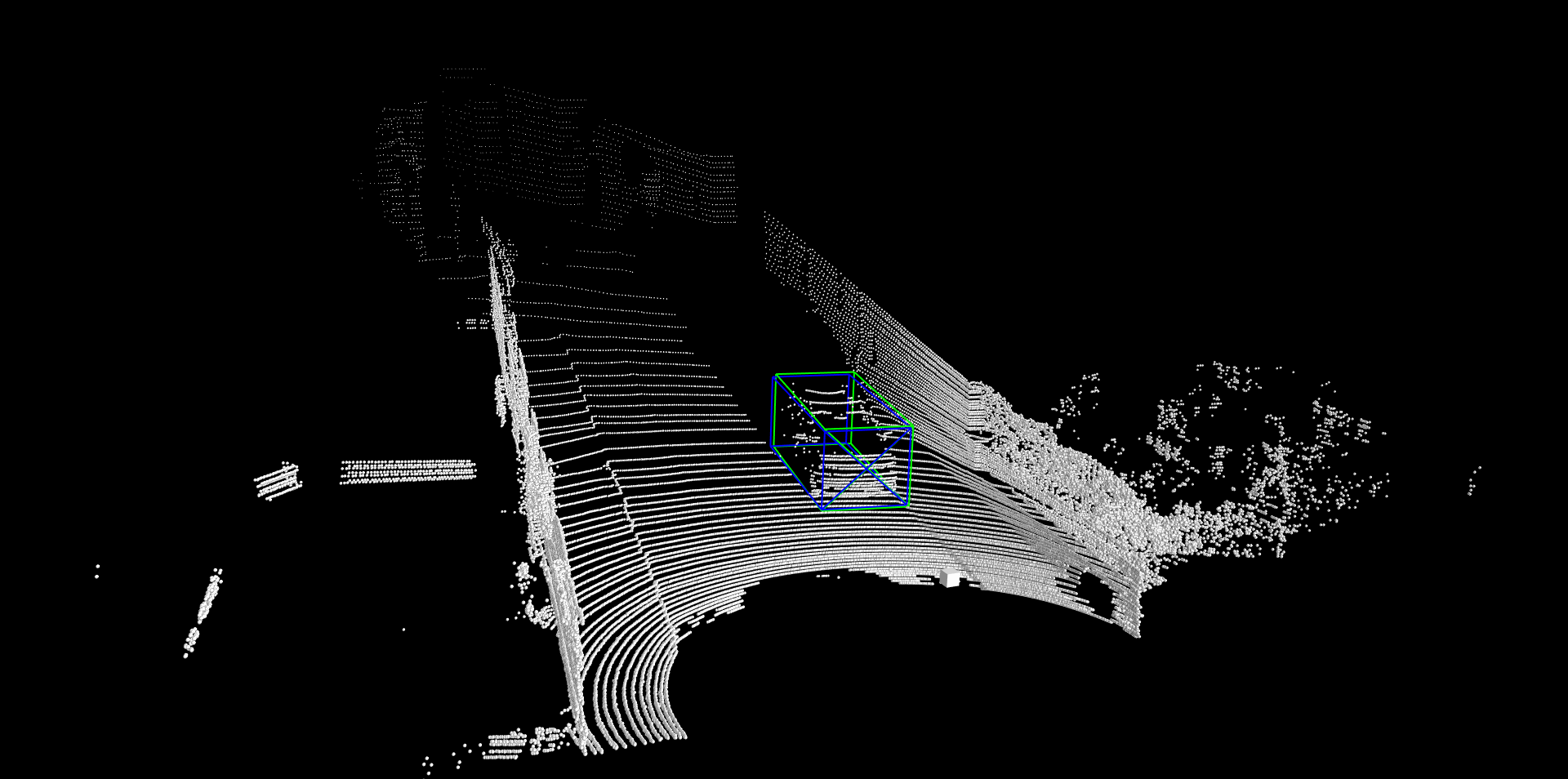}  
    \includegraphics[width=0.24\linewidth]{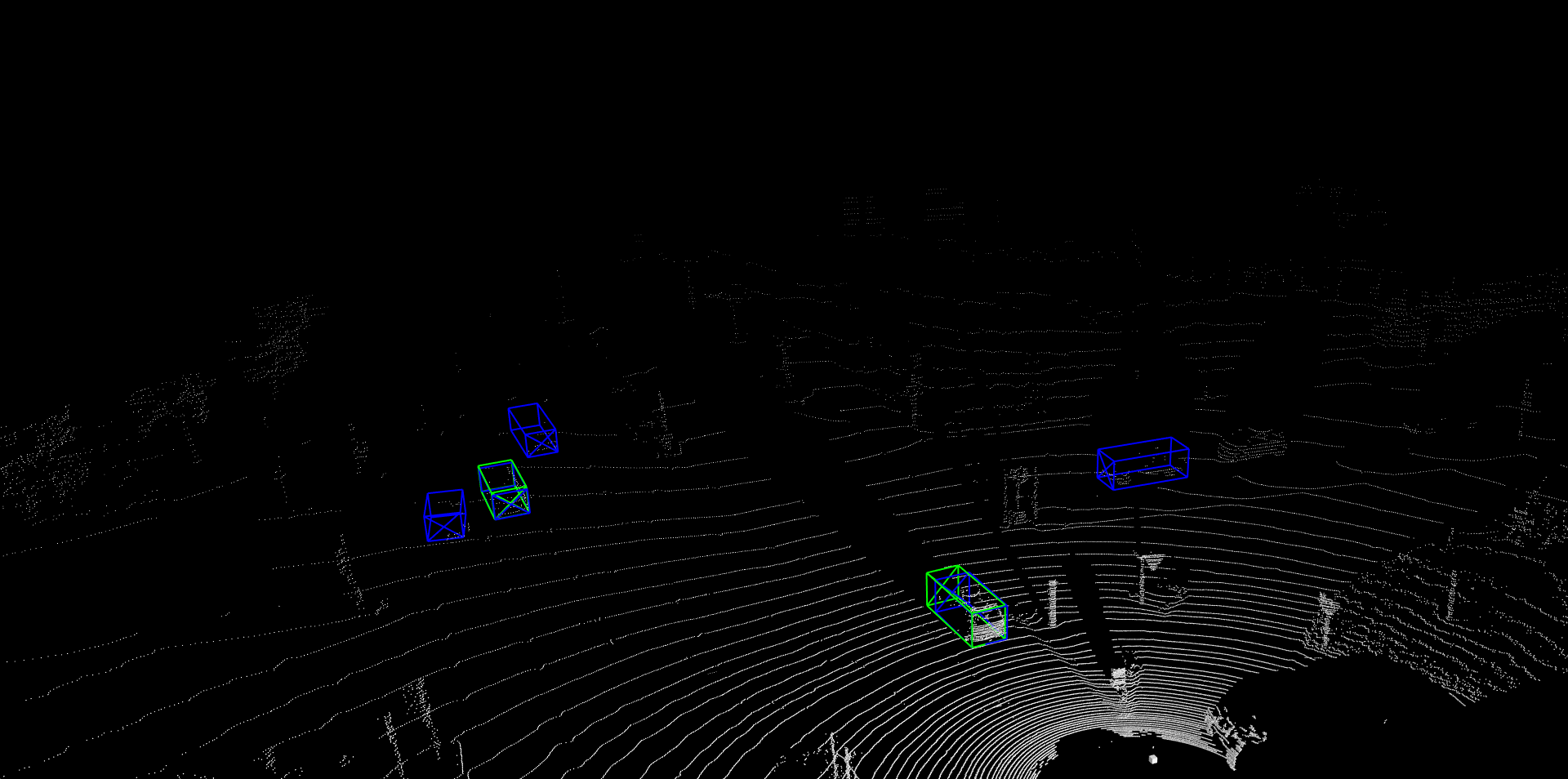} 
  \includegraphics[width=0.24\linewidth]{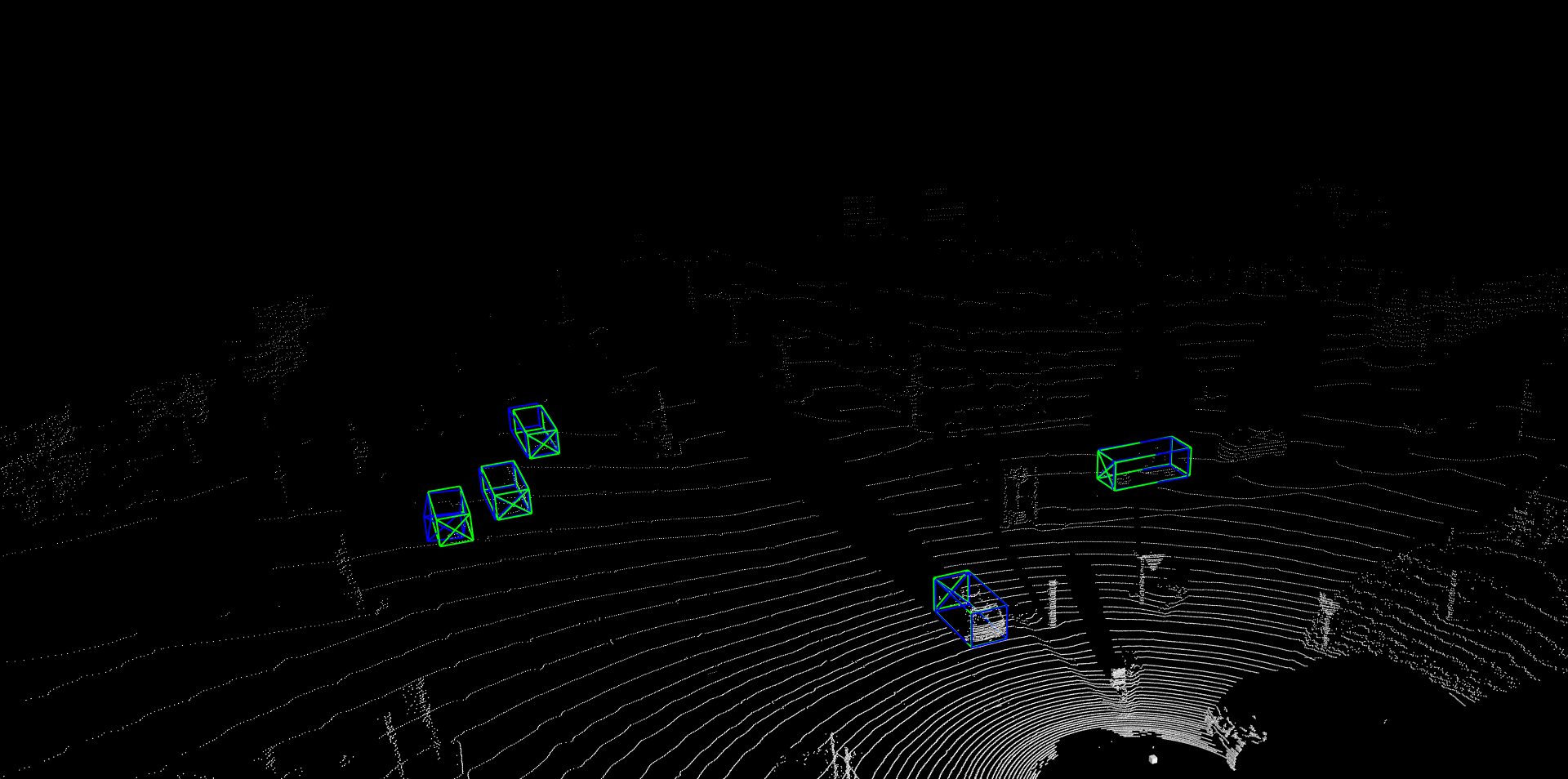}  
    \includegraphics[width=0.24\linewidth]{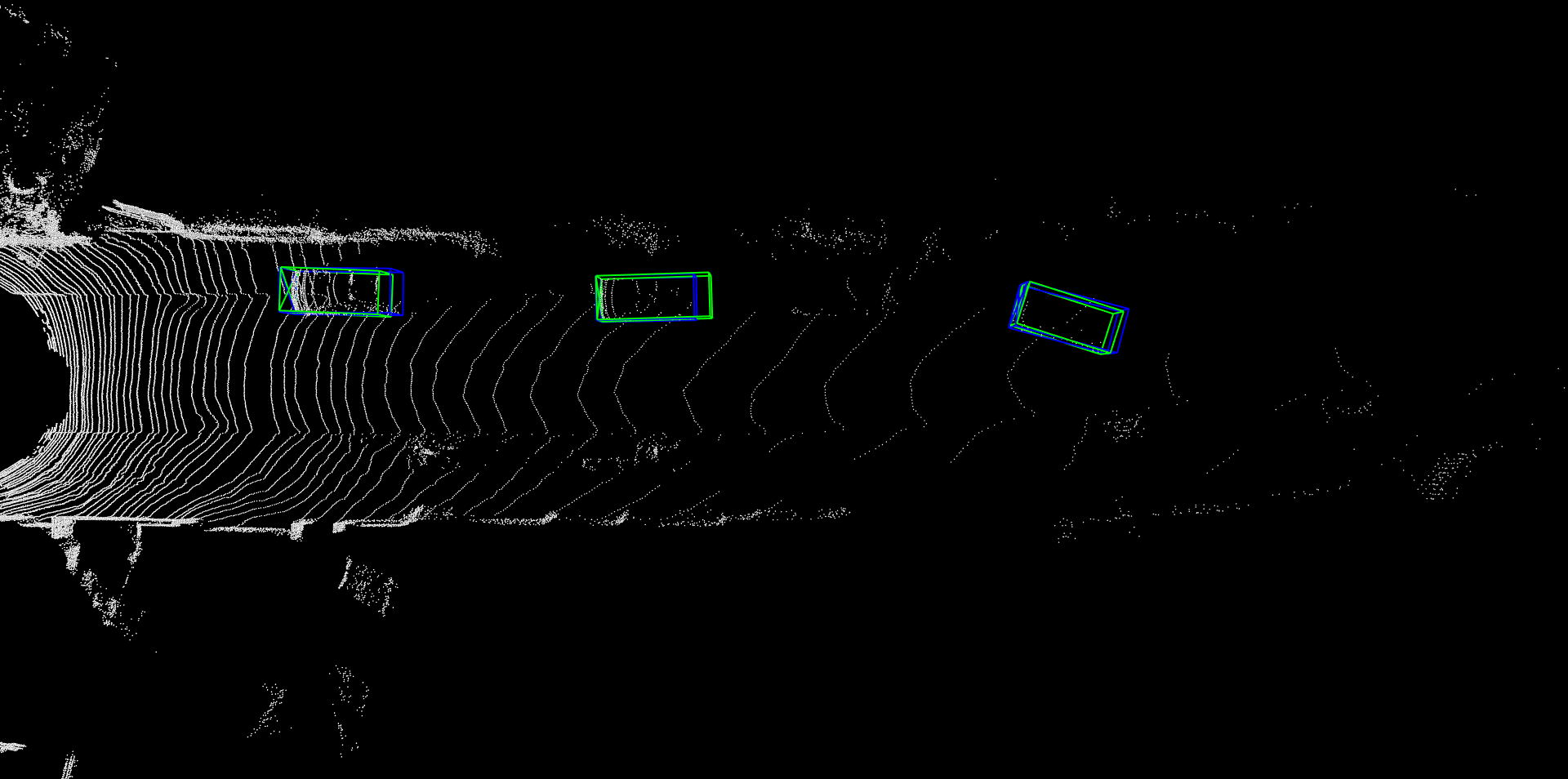} 
  \includegraphics[width=0.24\linewidth]{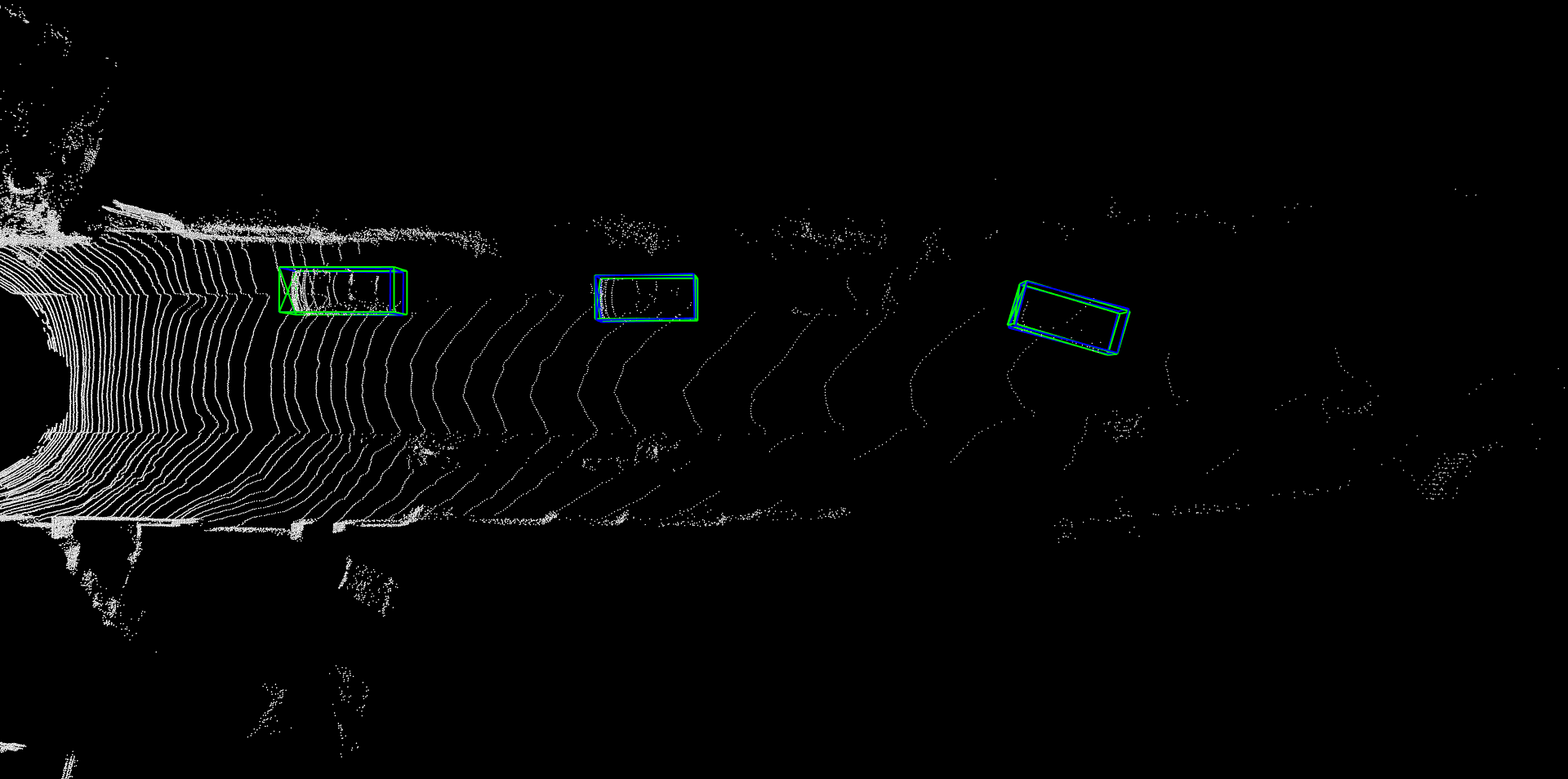}  
    \includegraphics[width=0.24\linewidth]{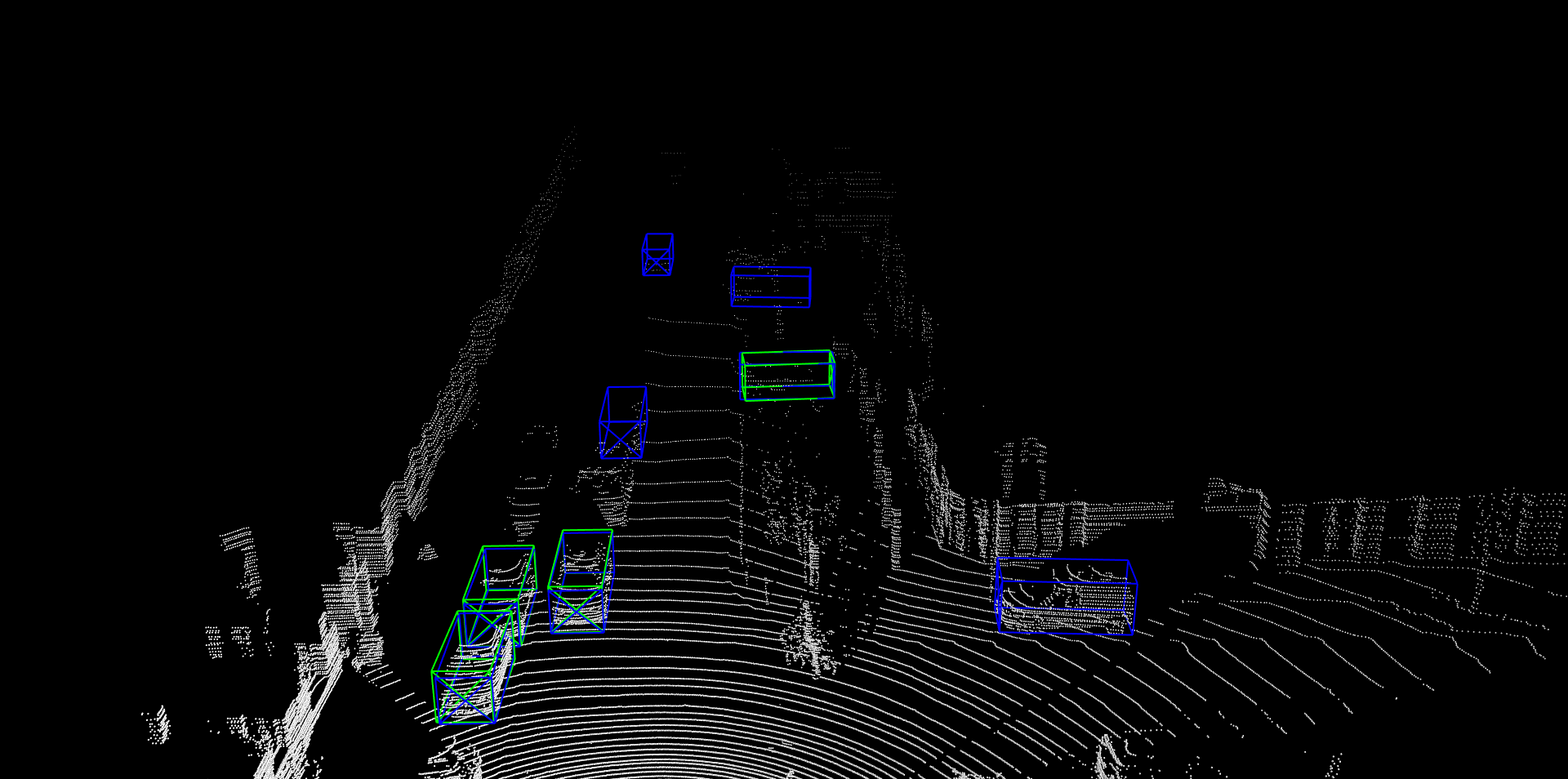} 
  \includegraphics[width=0.24\linewidth]{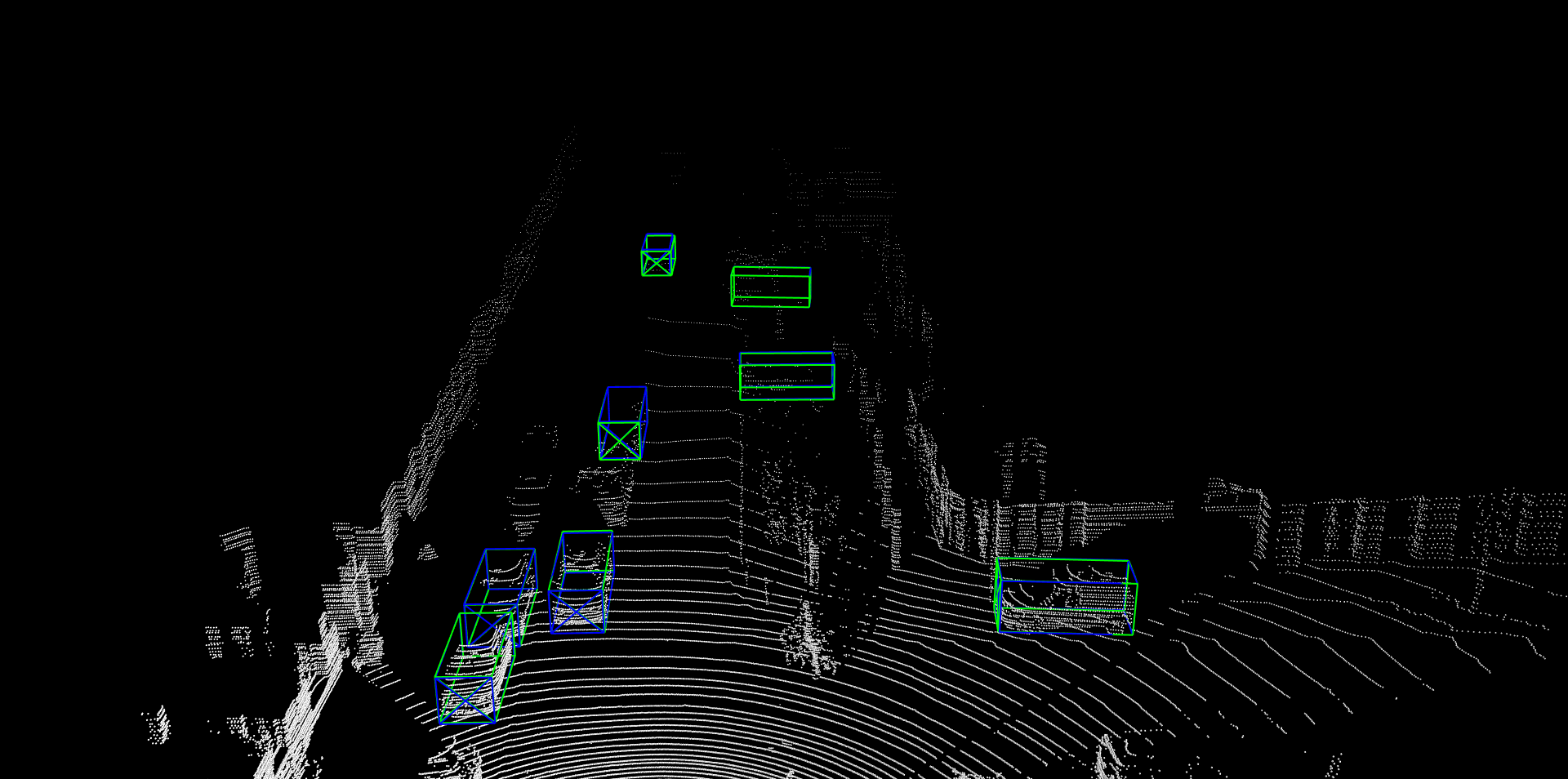}  
    \includegraphics[width=0.24\linewidth]{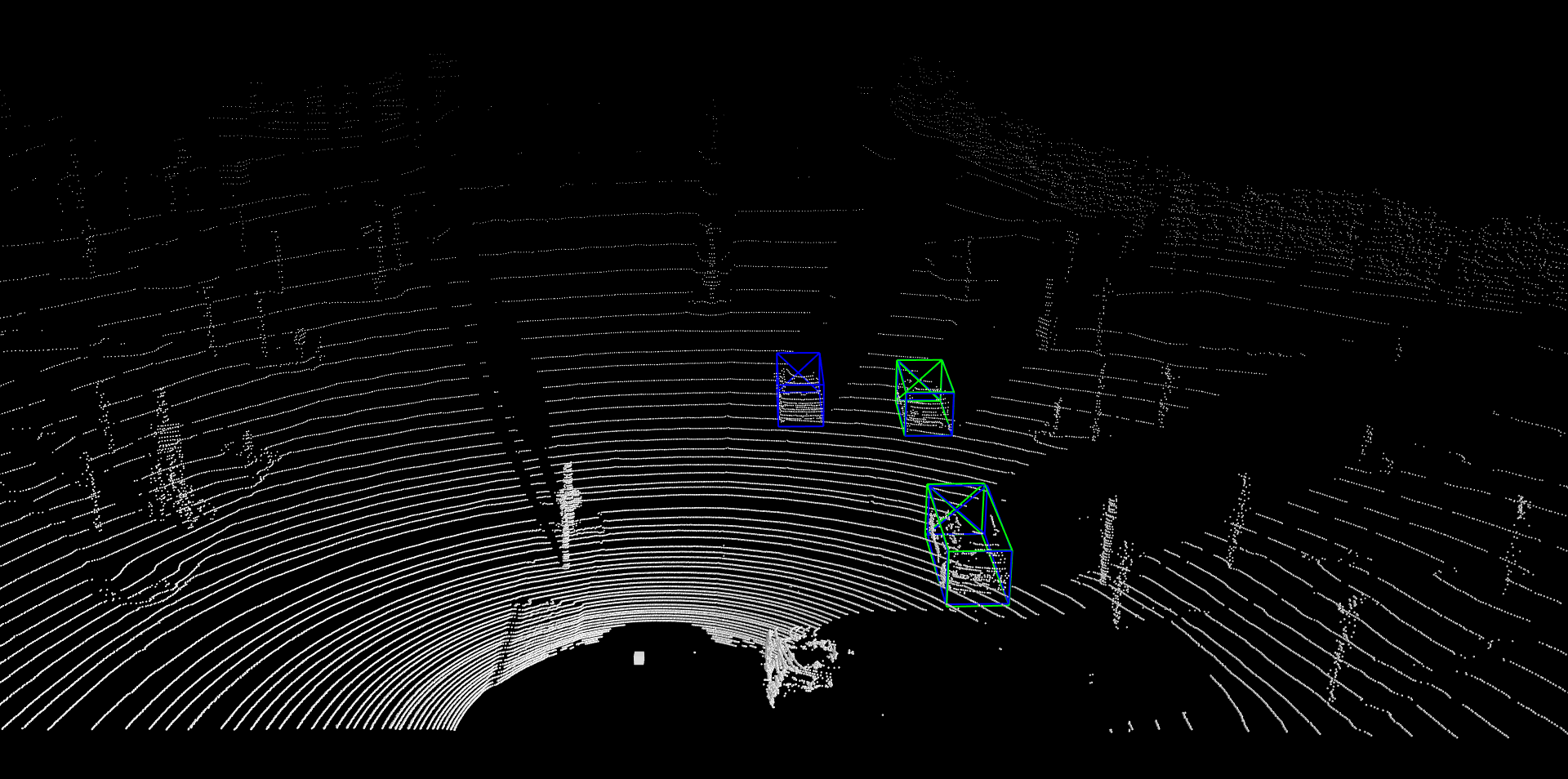} 
  \includegraphics[width=0.24\linewidth]{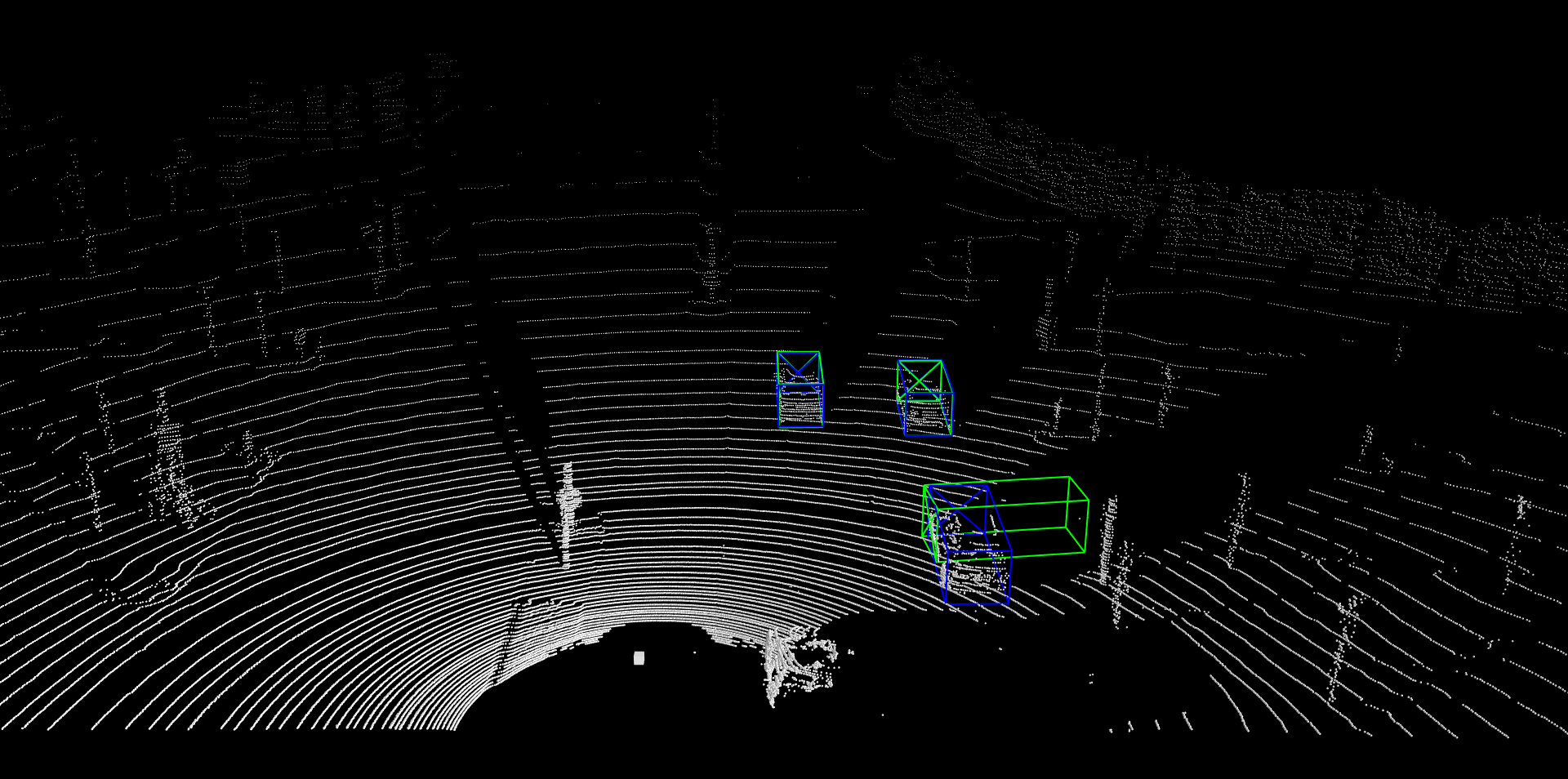}  
    \includegraphics[width=0.24\linewidth]{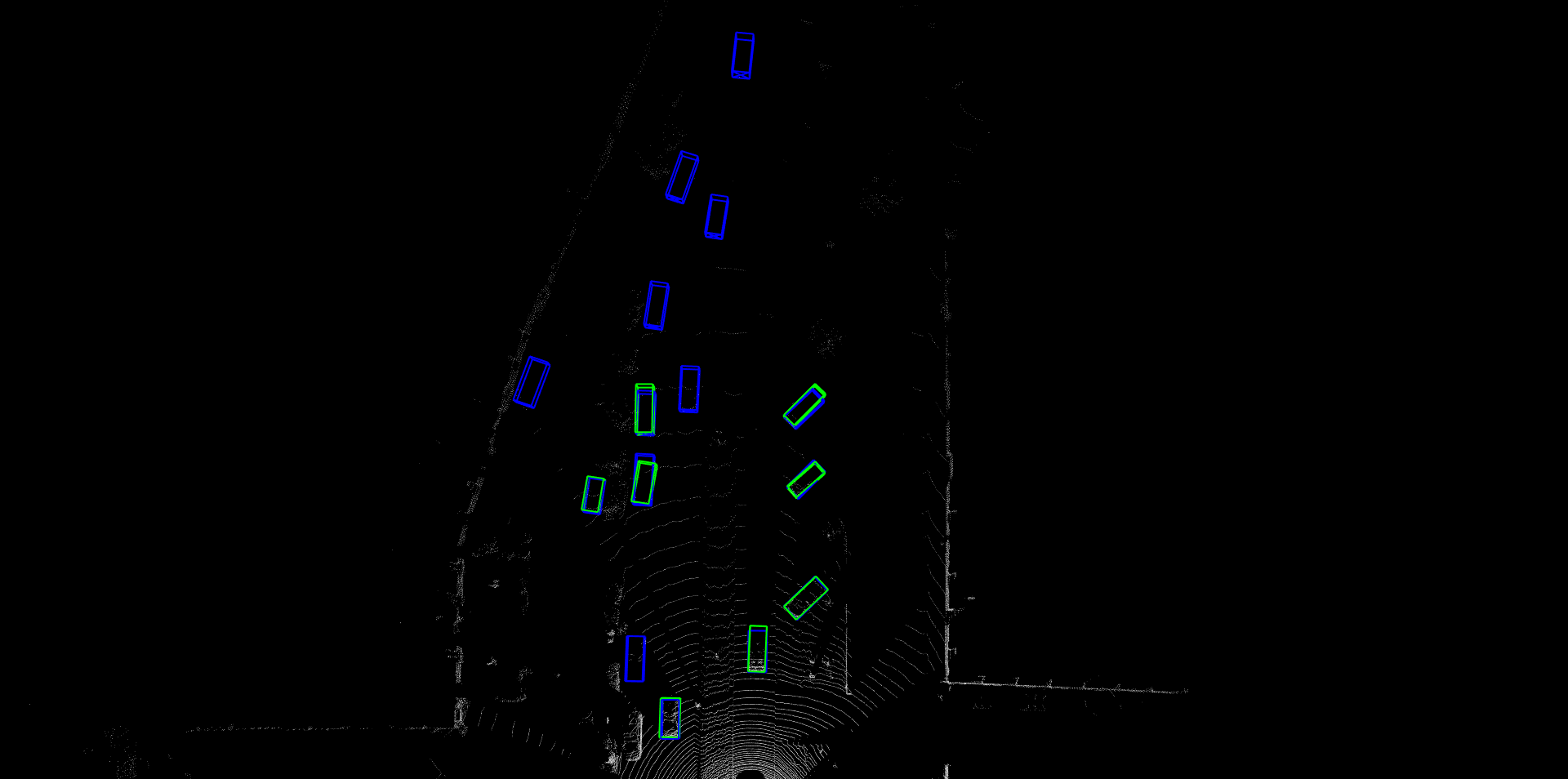} 
  \includegraphics[width=0.24\linewidth]{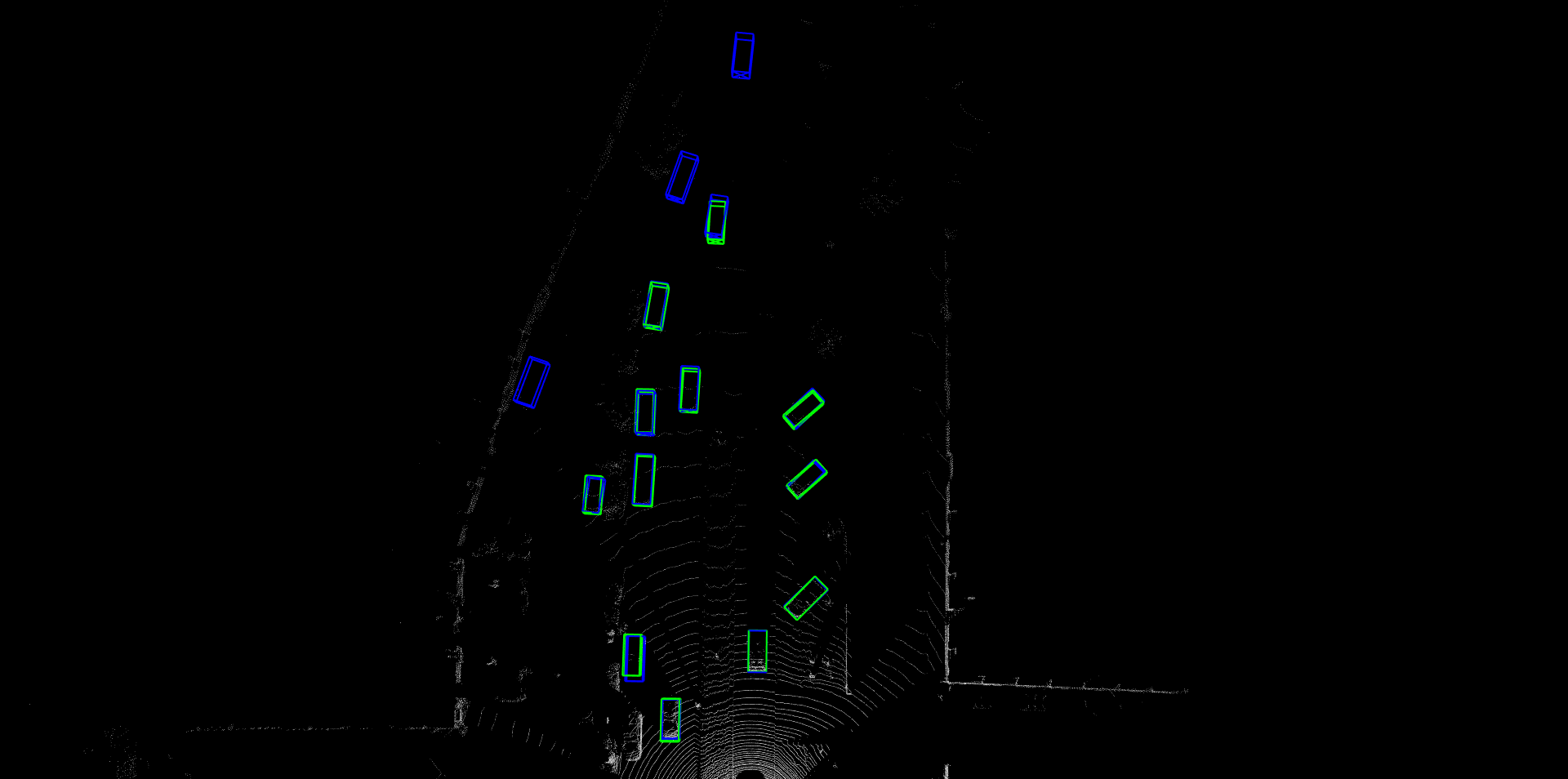}  
    \includegraphics[width=0.24\linewidth]{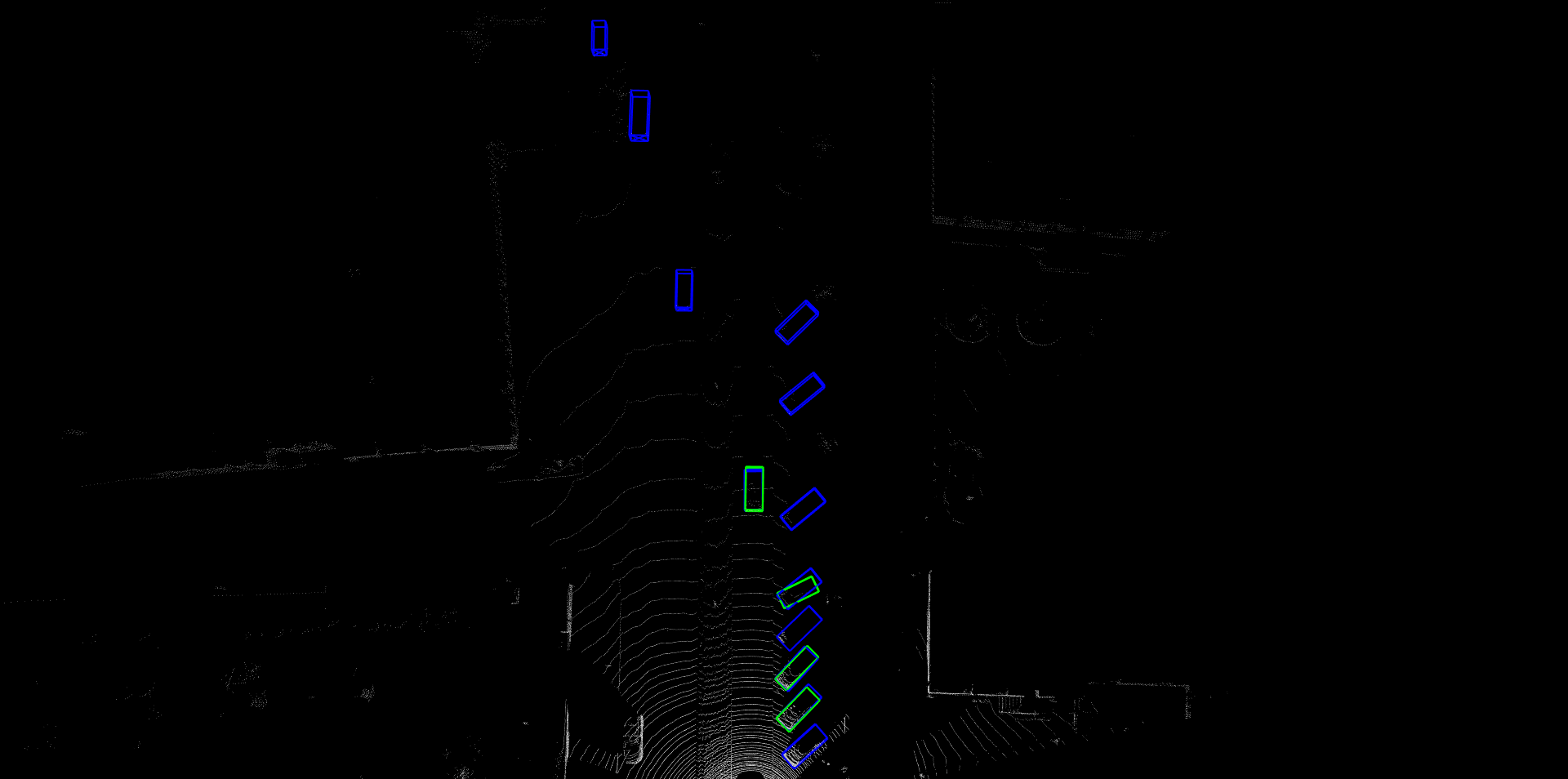} 
  \includegraphics[width=0.24\linewidth]{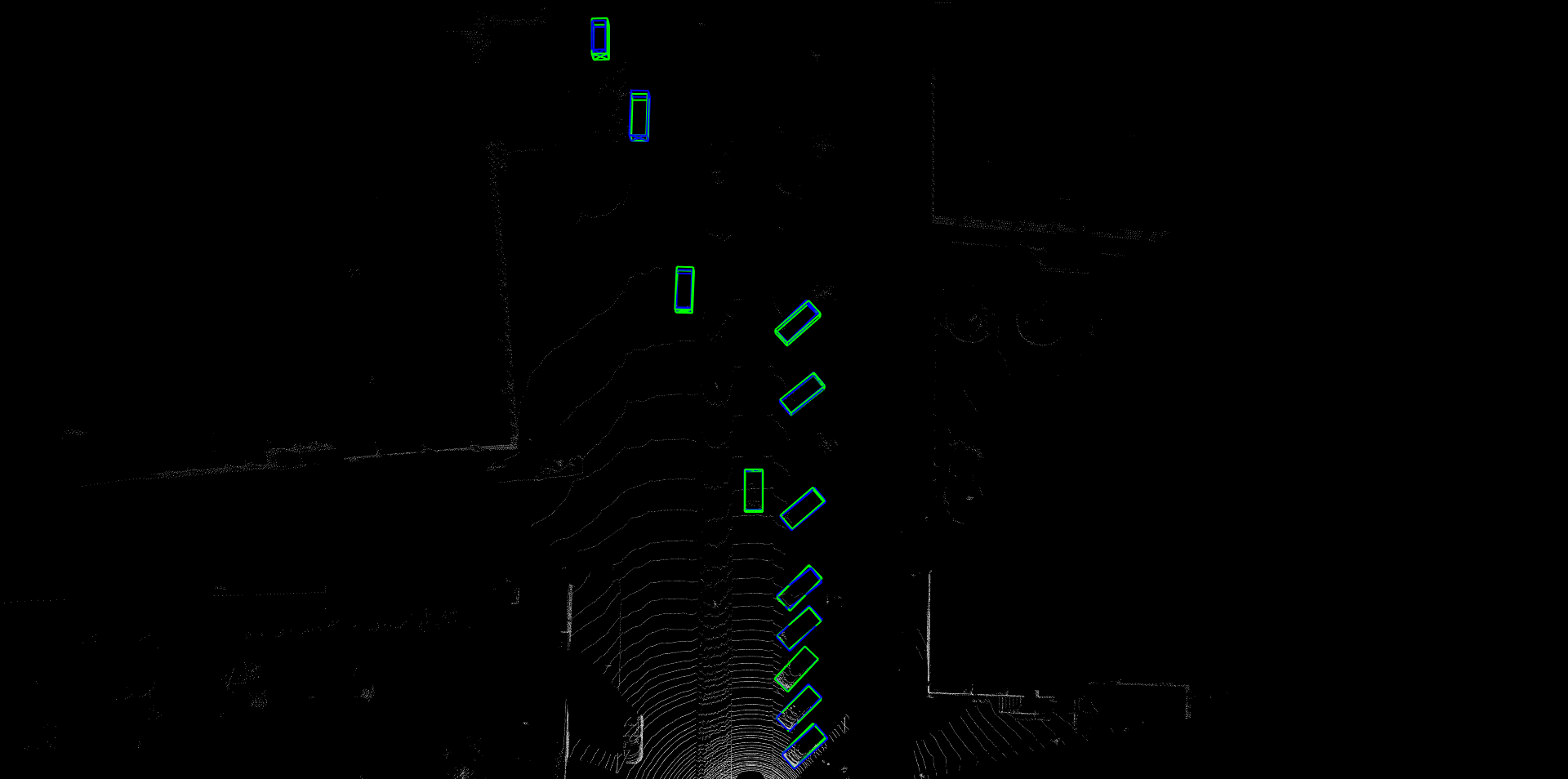}  
    \includegraphics[width=0.24\linewidth]{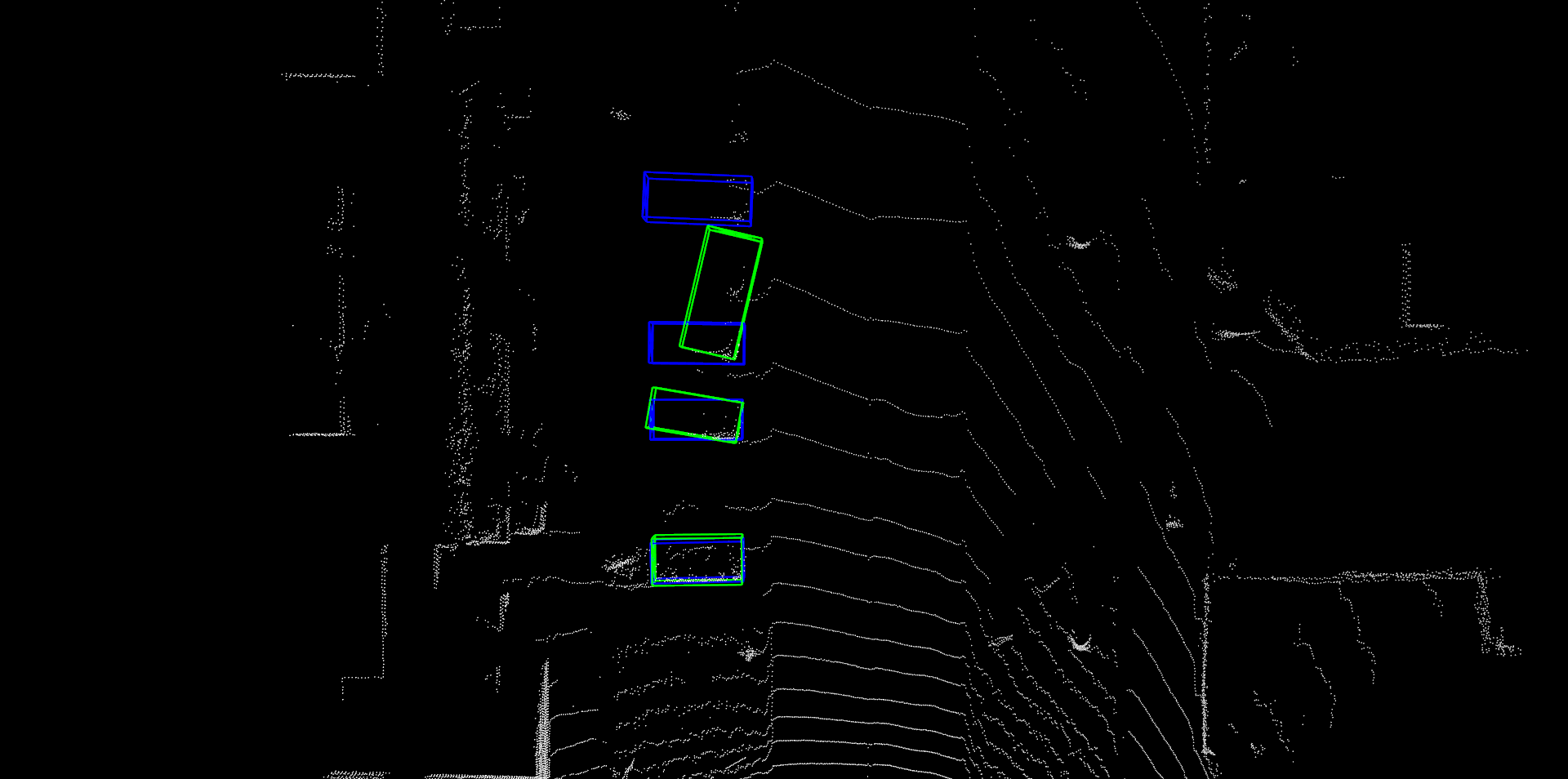} 
  \includegraphics[width=0.24\linewidth]{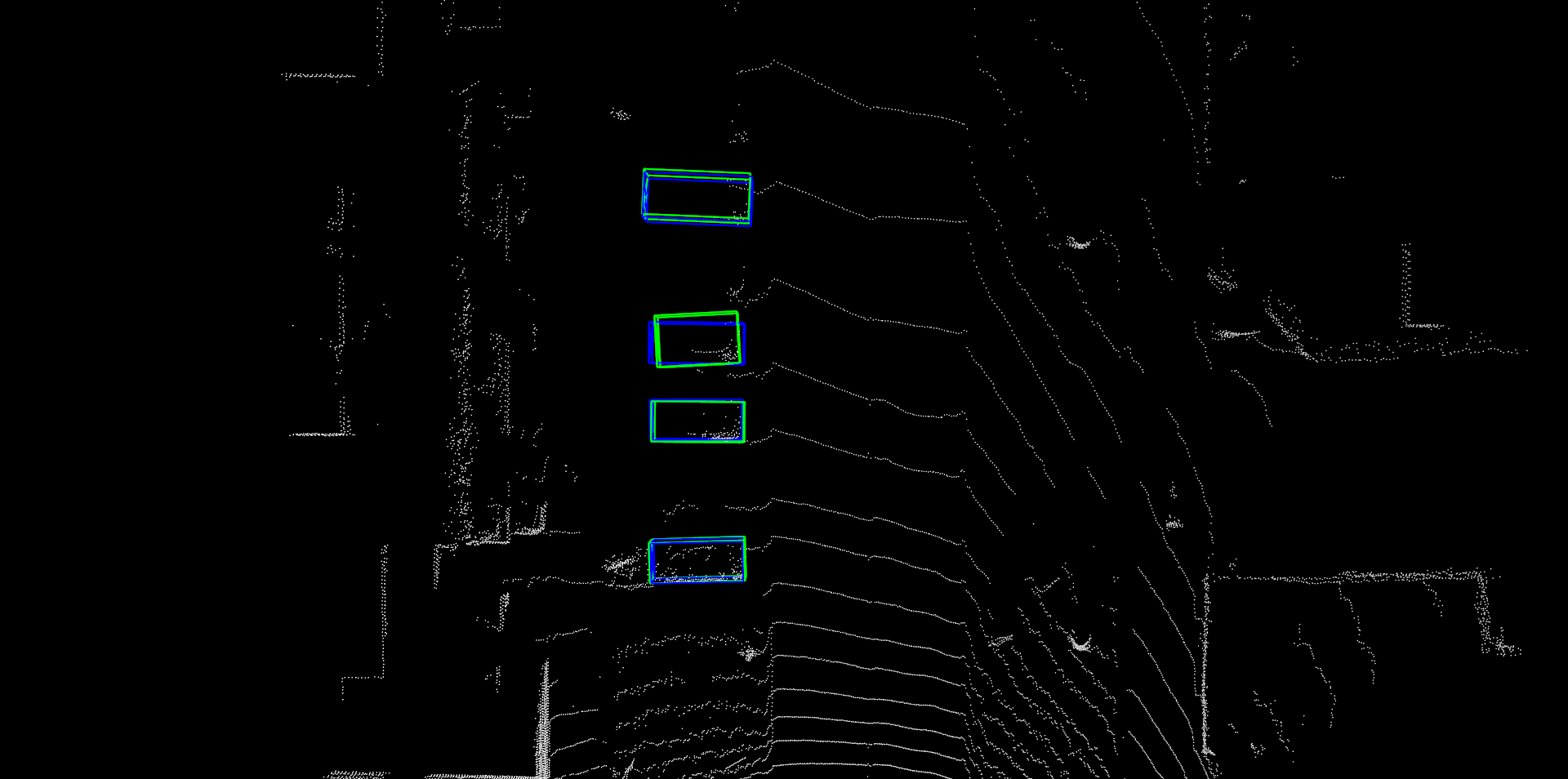}  
    \includegraphics[width=0.24\linewidth]{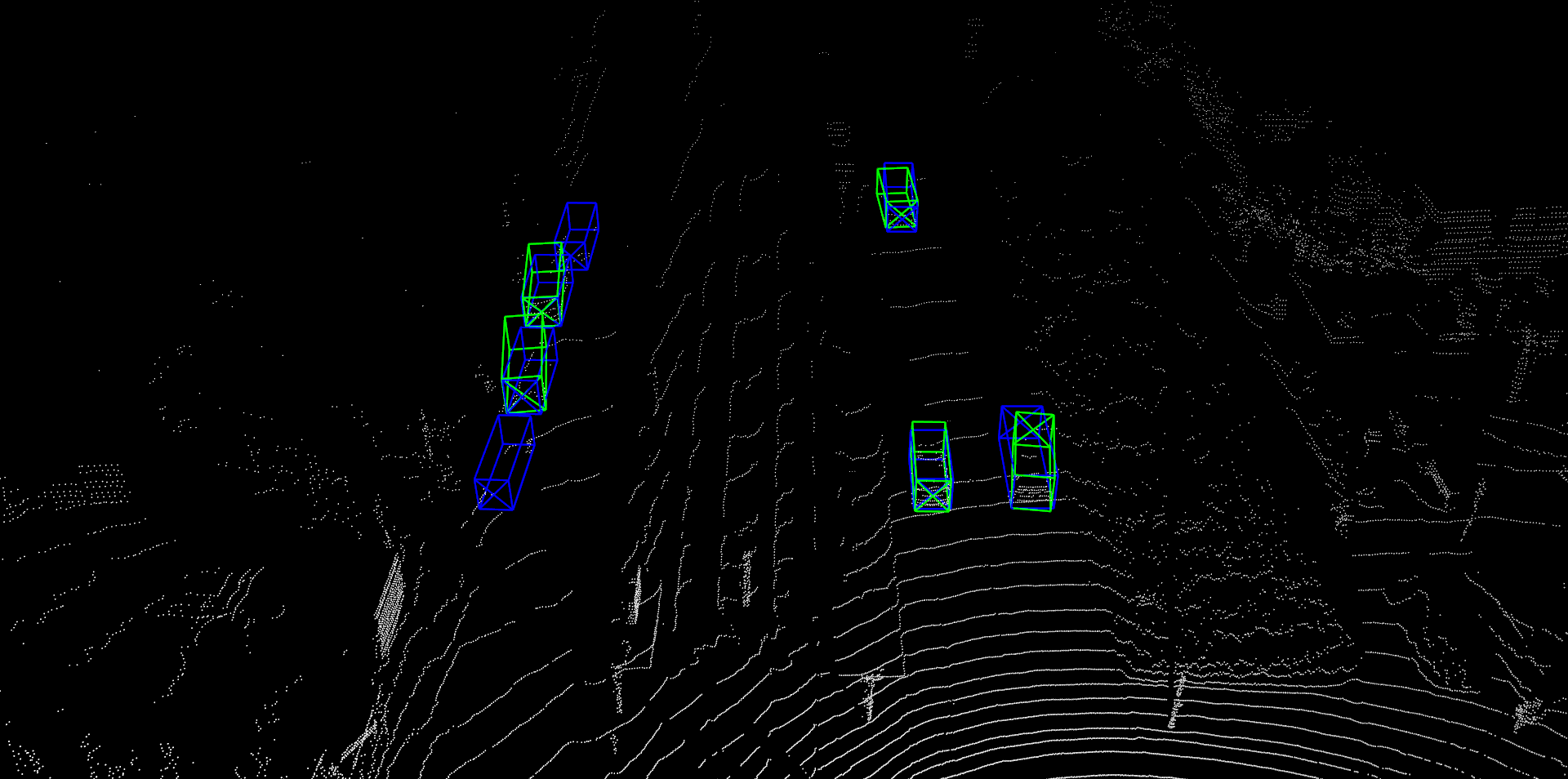} 
  \includegraphics[width=0.24\linewidth]{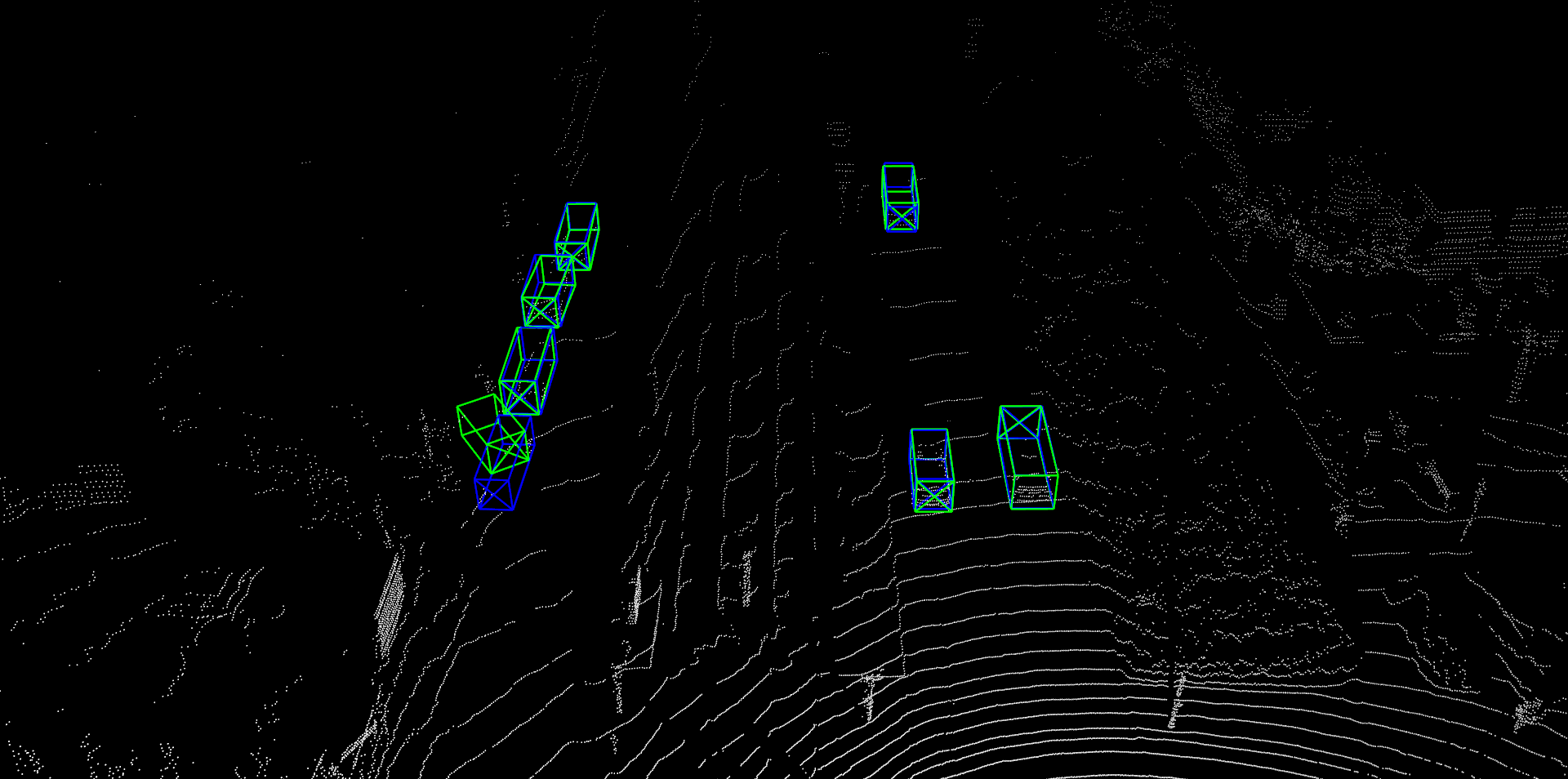}  
    \includegraphics[width=0.24\linewidth]{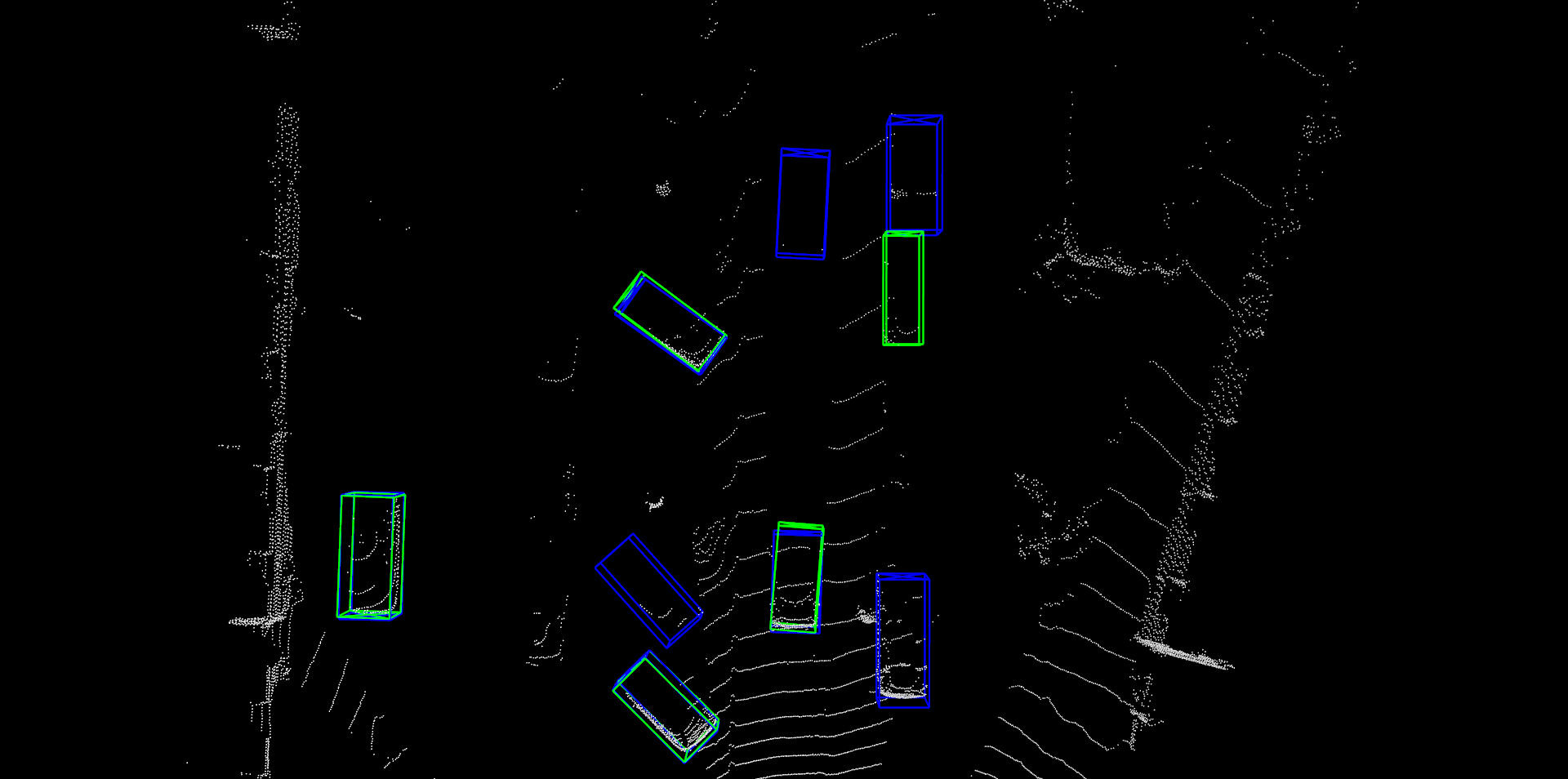} 
  \includegraphics[width=0.24\linewidth]{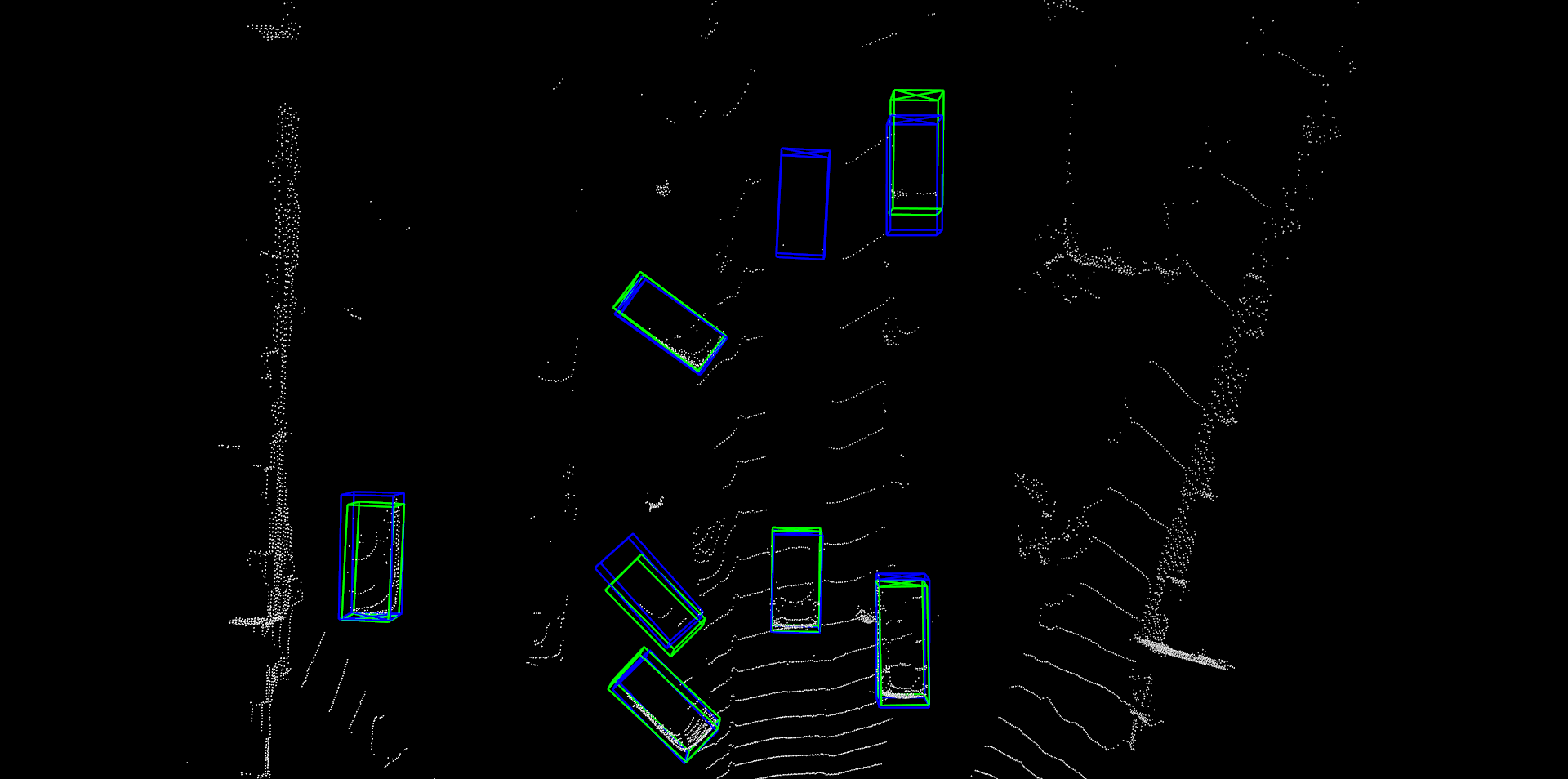}  
    \includegraphics[width=0.24\linewidth]{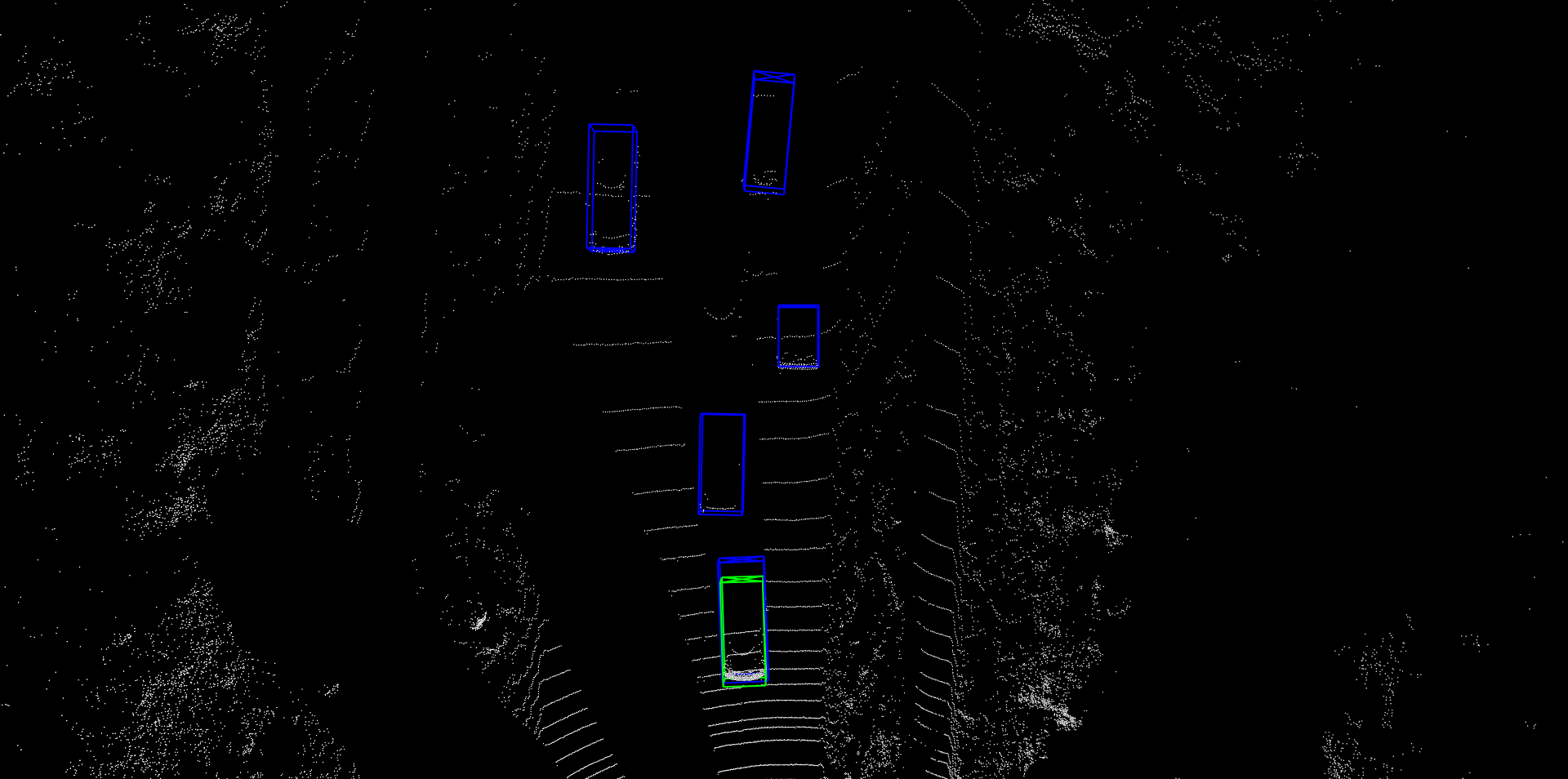} 
  \includegraphics[width=0.24\linewidth]{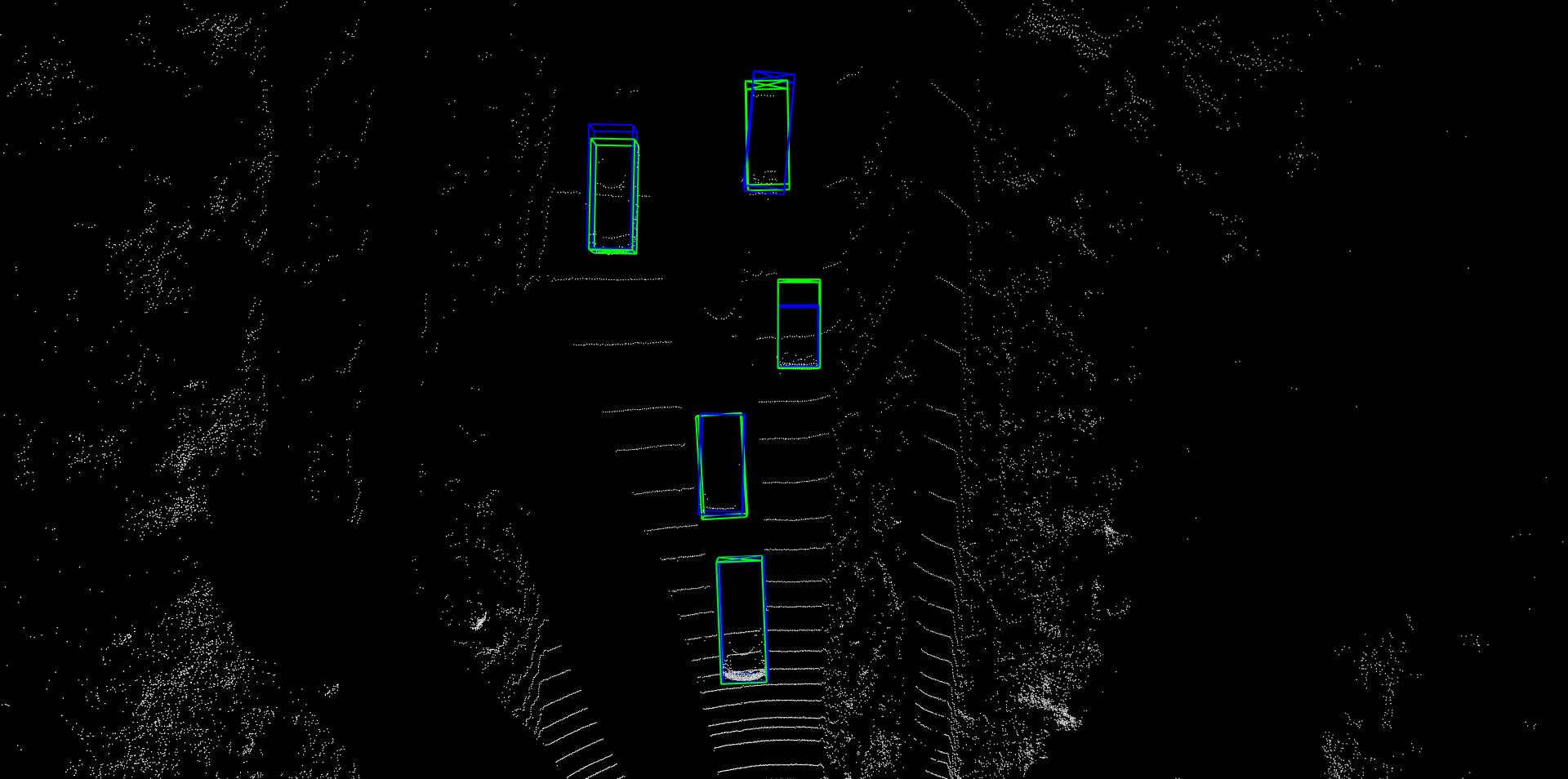}  
      \end{center}
   \caption{Comparison with the repeatable SOTA method, FGR. Each pair of the visualization contains results from FGR (left) and MTrans (Right). Blue: ground truth 3D box, Green: generated 3D box. Zoom in for a better view.}
   \label{fig:compare_with_FGR}
   \vspace{-6pt}
\end{figure*}

In \cref{fig:compare_with_FGR}, we compare the generated 3D boxes from our MTrans with the repeatable state-of-the-art baseline of FGR. As shown in the graph, FGR misses some objects and generates no box for them. It turns out that the objects containing fewer points (sparser) are more likely to be missed. Also, the FGR intentionally drops objects with low confidence, which even exacerbate this problem.

On the contrary, MTrans produces obviously higher quality 3D boxes in general. Compared with FGR, no objects are missed due to we force MTrans to generate one 3D box for each 2D weak annotation. However, some boxes are intentionally omitted because there are too few LiDAR points within the area. Compared with FGR, in some cases MTrans has some orientation errors (e.g., row3 \& row5 right in \cref{fig:compare_with_FGR}). We find those objects are positioned at the border of the 2D image and are truncated. Therefore incomplete frustum point clouds are fed into MTrans, which makes it difficult to predict 3D boxes.

\subsection{Visualization of Point Cloud Densification}
\begin{figure*}[t]
    \begin{center}
  \includegraphics[width=0.15\linewidth]{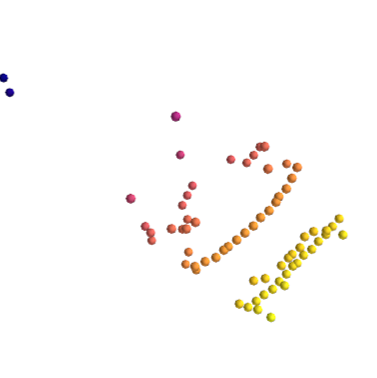} 
  \includegraphics[width=0.15\linewidth]{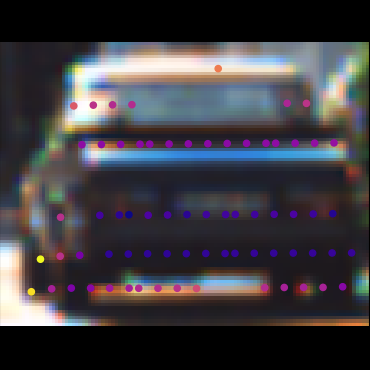}  
    \includegraphics[width=0.15\linewidth]{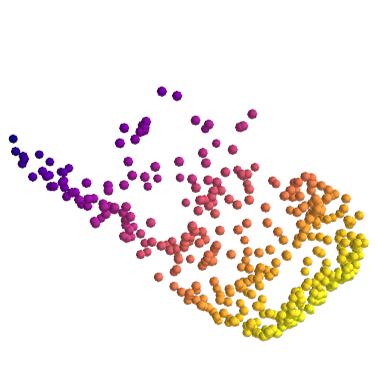} 
   \includegraphics[width=0.15\linewidth]{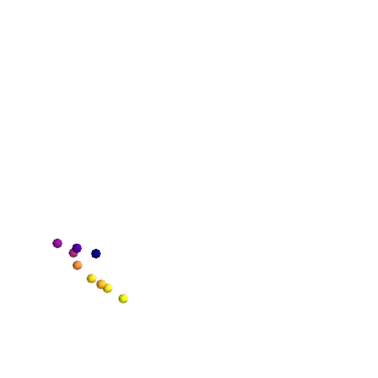} 
  \includegraphics[width=0.15\linewidth]{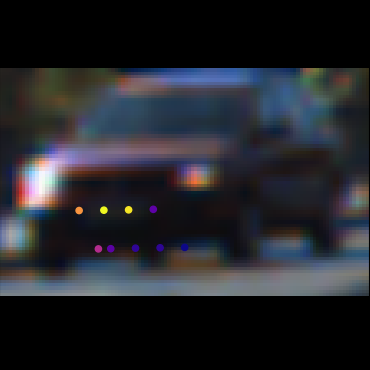}  
    \includegraphics[width=0.15\linewidth]{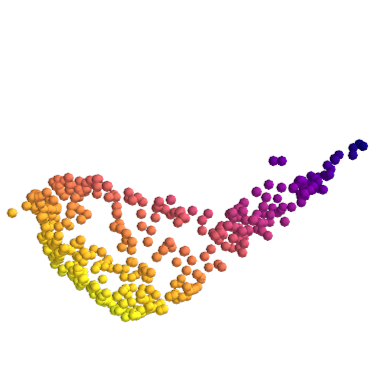}   \includegraphics[width=0.15\linewidth]{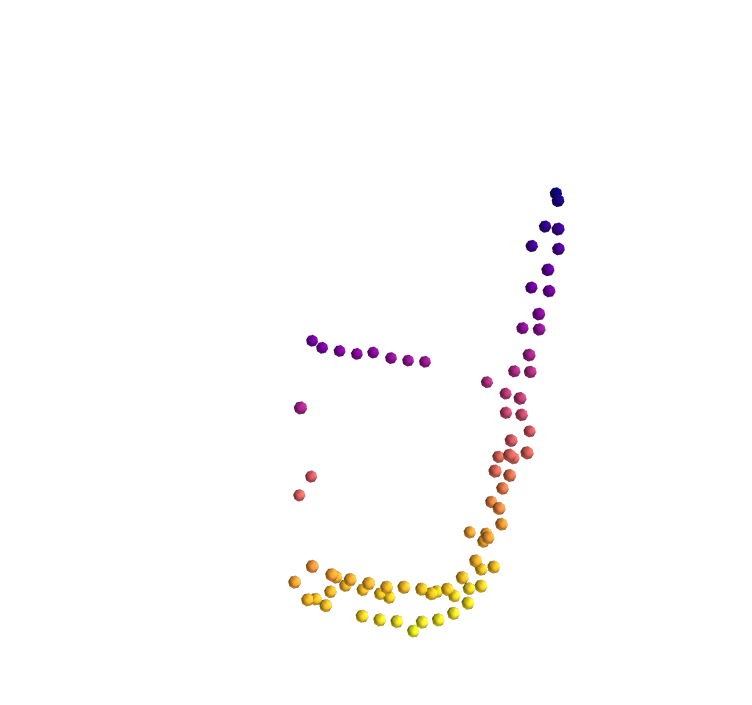} 
  \includegraphics[width=0.15\linewidth]{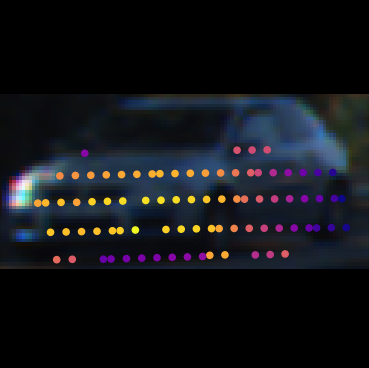}  
    \includegraphics[width=0.15\linewidth]{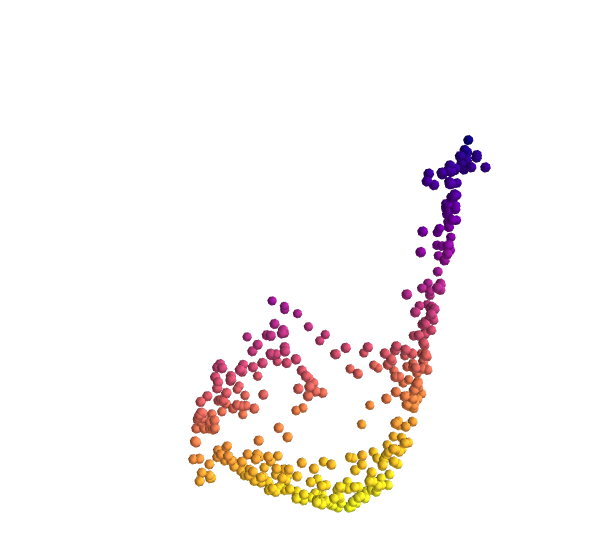}   \includegraphics[width=0.15\linewidth]{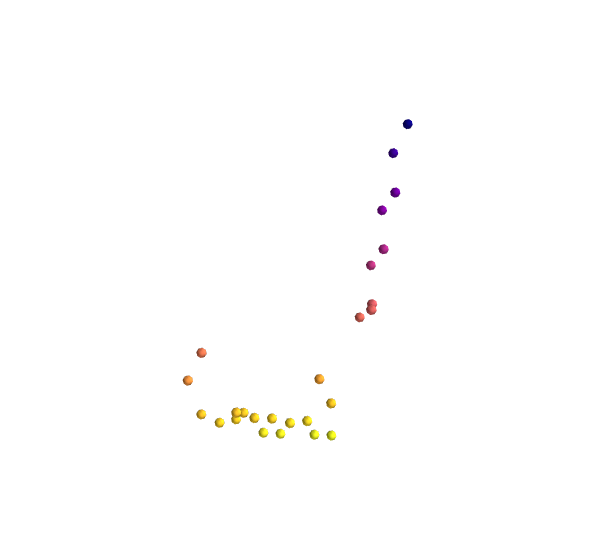} 
  \includegraphics[width=0.15\linewidth]{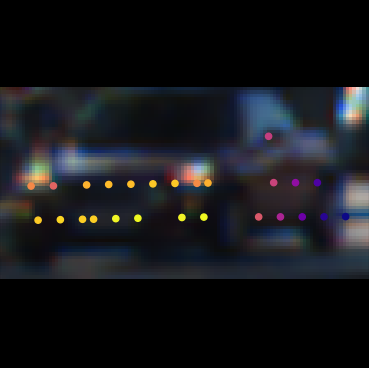}  
    \includegraphics[width=0.15\linewidth]{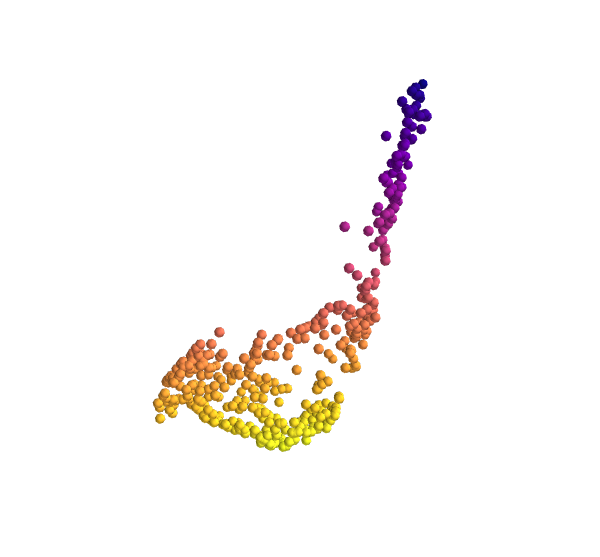}   \includegraphics[width=0.15\linewidth]{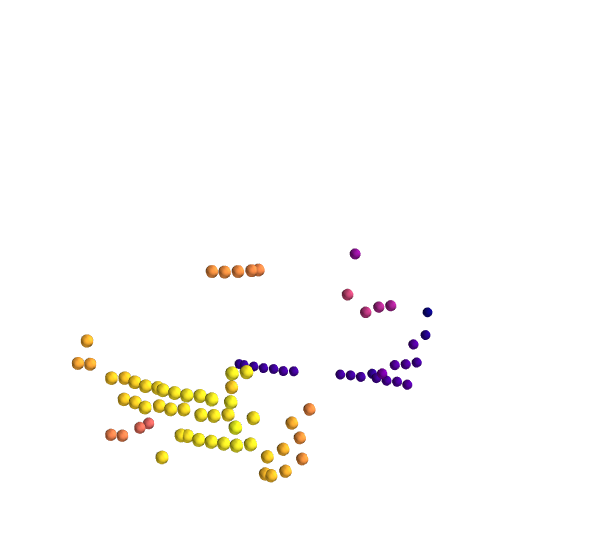} 
  \includegraphics[width=0.15\linewidth]{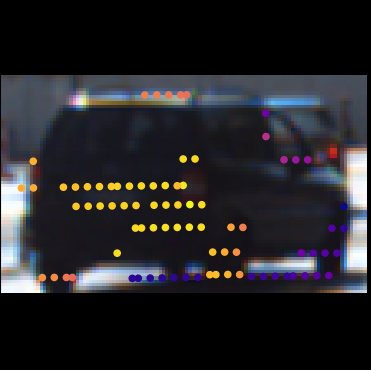}  
    \includegraphics[width=0.15\linewidth]{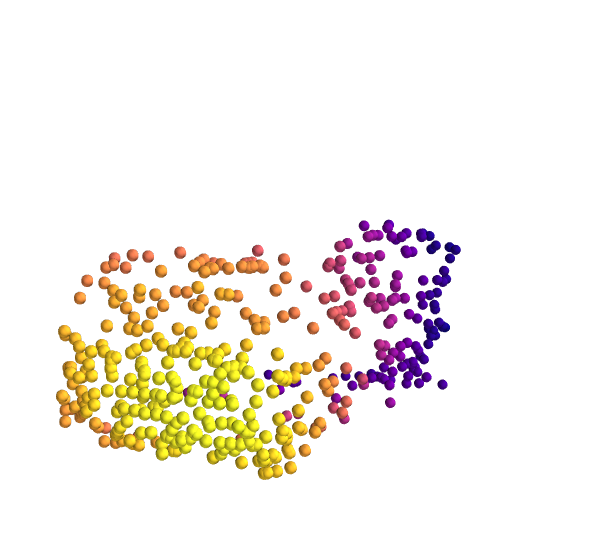}   \includegraphics[width=0.15\linewidth]{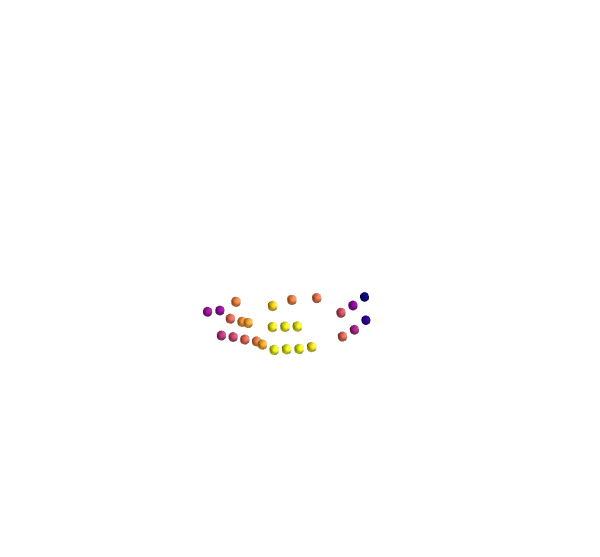} 
  \includegraphics[width=0.15\linewidth]{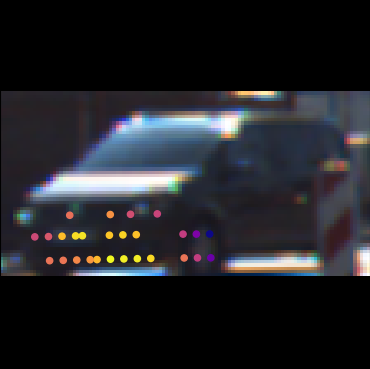}  
    \includegraphics[width=0.15\linewidth]{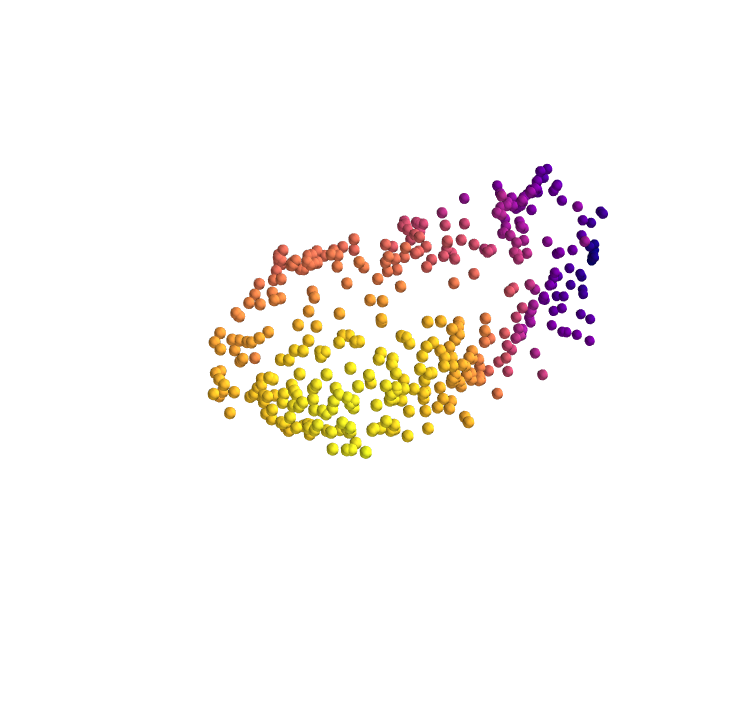} 
      \end{center}
   \caption{Qualitative results of the densification of the point clouds. Each row contains the original sparse point cloud (left), the corresponding image with LiDAR points projected (middle) and the densified point cloud (right). Distances of the LiDAR points are denoted by the color.}
   \label{fig:densify_visualization}
   \vspace{-6pt}
\end{figure*}

In this section, we visualize the densified point clouds to justify the effectiveness of MTrans. Given a sparse point cloud and the corresponding image, MTrans performs image-guided interpolation, estimating 3D coordinates for image pixels. By doing so, extra 3D points are generated and hence the original sparse point cloud is densified. As shown in \cref{fig:densify_visualization}, the sparse point clouds are densified in high fidelity. The distribution of the generated points follows the object shape depicted in the image. Especially for the extreme cases (row1 \& row3 right), the car shapes are recovered from the original clouds that are hardly recognizable due to the sparsity. The visualization results prove the effectiveness of the deification procedure of our MTrans, which leverages the 2D image information to enrich the sparse point clouds.

\clearpage
%
%
\bibliographystyle{splncs04}
\bibliography{ref.bib}